\documentclass[conference]{IEEEtran}
\IEEEoverridecommandlockouts
\usepackage{cite}
\usepackage{amsmath,amssymb,amsfonts}
\usepackage{algorithmic}
\usepackage{graphicx}
\usepackage{textcomp}
\usepackage{xcolor}
\usepackage{subfig}
\usepackage{hyperref} 
\usepackage{array}
\def\BibTeX{{\rm B\kern-.05em{\sc i\kern-.025em b}\kern-.08em
    T\kern-.1667em\lower.7ex\hbox{E}\kern-.125emX}}
\begin{document}
\title{An Evaluation of Standard Statistical Models and LLMs on Time Series Forecasting
\thanks{979-8-3503-7565-7/24/\$31.00 ©2024 IEEE}
}

\author{
\IEEEauthorblockN{Rui Cao}
\IEEEauthorblockA{\textit{School of Information Science and Engineering} \\
\textit{Southeast University}\\
Nanjing, China \\
\href{mailto:ruicao@seu.edu.cn}{ruicao@seu.edu.cn}}
\and
\IEEEauthorblockN{Qiao Wang}
\IEEEauthorblockA{\textit{School of Information Science and Engineering}\\
\textit{and}\\
\textit{School of Economics and Management} \\
\textit{Southeast University}\\
Nanjing, China \\
\href{mailto:qiaowang@seu.edu.cn}{qiaowang@seu.edu.cn}\ \ (Corresponding Author)}
\and

% \IEEEauthorblockN{ Given Name Surname}
% \and
% \IEEEauthorblockN{ Given Name Surname}
}
\maketitle
\begin{abstract}
This research examines the use of Large Language Models (LLMs) in predicting time series, with a specific focus on the LLMTIME model. Despite the established effectiveness of LLMs in tasks such as text generation, language translation, and sentiment analysis, this study highlights the key challenges that large language models encounter in the context of time series prediction. We assess the performance of LLMTIME across multiple datasets and introduce classical almost periodic functions as time series to gauge its effectiveness. The empirical results indicate that while large language models can perform well in zero-shot forecasting for certain datasets, their predictive accuracy diminishes notably when confronted with diverse time series data and traditional signals. The primary finding of this study is that the predictive capacity of LLMTIME, similar to other LLMs, significantly deteriorates when dealing with time series data that contain both periodic and trend components, as well as when the signal comprises complex frequency components. %{https://github.com/crSEU/llmtimeVSarima}.

\end{abstract}
\section{Introduction}
Time series analysis is a highly practical and fundamental issue that holds significant importance in various real-world scenarios \cite{b34}\cite{b35}, such as predicting retail sales as discussed by Böse et al. \cite{b1}, filling in missing data in economic time series according to Friedman \cite{b2}, identifying anomalies in industrial maintenance per Gao et al. \cite{b3}, and categorizing time series data from different domains by Ismail Fawaz et al. \cite{b4}. Numerous statistical and machine learning techniques have been created over time for time series analysis. Drawing inspiration from the remarkable success seen in natural language processing and computer vision, transformers \cite{b21} have been integrated into various tasks involving time series data. Wen et al. \cite{b6} demonstrated promising outcomes, particularly in time series forecasting as shown by Lim et al. \cite{b7}; Nie et al. \cite{b8}.

% Time series modeling is similar in many ways to other sequence modeling problems, such as text, audio, or video, but does present two uniquely challenging problems. Firstly, unlike video and audio signals, which typically have consistent input ratios and sampling rates,

Time series datasets frequently consist of sequences originating from various sources, each potentially characterized by distinct time scales, durations, and sampling frequencies. This diversity poses difficulties in both model training and data processing. Moreover, missing values are common in time series data, necessitating specific approaches to address these gaps to maintain the precision and resilience of the model. Furthermore, prevalent applications of time series forecasting, such as weather prediction or financial analysis, entail making predictions based on observations with limited available information, rendering forecasting a challenging task. Ensuring both accuracy and reliability is complex due to the necessity of handling unknown and unpredictable variables when extrapolating future data, underscoring the importance of estimating uncertainty. Consequently, it is crucial for the model to possess strong generalization capabilities and an effective mechanism for managing uncertainty to adeptly adapt to forthcoming data variations and obstacles.

Currently, large-scale pre-training has emerged as a crucial technique for training extensive neural networks in the domains of vision and text, leading to a significant enhancement in performance \cite{b9}\cite{b10}. Nonetheless, the utilization of pre-training in time series modeling faces certain challenges. Time series data, unlike visual and textual data, lacks well-defined unsupervised targets, posing difficulties in achieving effective pre-training. Moreover, the scarcity of large-scale and coherent pre-trained datasets for time series further hinders the adoption of pre-training in this domain. Consequently, conventional time-series approaches (e.g., ARIMA \cite{b11} and linear models \cite{b12}) often outperform deep learning methods in widely used benchmarks \cite{b13}. With the advent of pre-trained models, many researchers have started to forecast time series by leveraging LLMs, such as the LLMTIME technique \cite{b14}, illustrating how LLMs naturally bridge the gap between the simplistic biases of traditional methods and the intricate representational learning and generation capabilities of contemporary deep learning. The pre-trained LLMs are applied to tackle the challenge of continuous time series forecasting \cite{b32}\cite{b33}. However, while this method demonstrates comparable performance to traditional processing methods on some datasets, extensive experimentation in this study reveals that the LLMTIME approach lacks the ability for zero-shot time series forecasting and even underperforms compared to the traditional time series method ARIMA when tested on a diverse set of datasets.

\section{Background}

\subsection{Language Model}

A language model is created to determine the probability that a specific sequence of words forms a coherent sentence. For example, let's consider two instances: Sequence A = {never, too, late, to, learn}, which clearly constructs a sentence ("never too late to learn."), and a proficient language model should assign it a high likelihood. Conversely, the word sequence B = {ever, zoo, later, too, eat} is less probable to create a coherent sentence, and a well-trained language model would assign it a lower probability. The main objectives of language models involve evaluating the probability of a word sequence forming a grammatically correct sentence and forecasting the probability of the next word based on the preceding words. To accomplish these objectives, language models work under the fundamental assumption that the occurrence of a subsequent word is influenced only by the words that come before it. This assumption allows the sentence's joint probability to be related to the conditional probability of each word in the sequence. Language models are trained on a set of sequences denoted as $U ={U_1, U_2, \ldots U_i, \ldots , U_N }$, where each sequence is depicted as $U_i = (u_1, u_2, \ldots , u_j , \ldots, u_{n_i})$, and each token $u_j$ ($j$ in $\{1, 2, \ldots , n_i \}$) belongs to a vocabulary $\mathcal{V}$. Large language models typically function as autoregressive distributions, implying that the probability of each token is dependent solely on the preceding tokens in the sequence, and can be formulated as $\begin{aligned}p_\theta\left(U_i\right)=\prod_{j=1}^{n_i}p_\theta\left(u_j\mid u_{0:j-1}\right)\end{aligned}$. In this context, the model parameters $\theta$ are acquired by maximizing the probability of the complete dataset, expressed as $\begin{aligned}p_\theta(\mathcal{U})=\prod_{i=1}^Np_\theta(U_i)\end{aligned}$.

\subsection{Large Language Model}
LLMs represent a type of artificial intelligence model created to comprehend and produce human language. They undergo training on extensive text datasets and have the capability to carry out various tasks, such as text creation \cite{b15,b16,b17}, machine translation \cite{b18}, dialogue systems \cite{b19}, among others. These models are distinguished by their immense size, encompassing billions of parameters that facilitate their understanding of intricate patterns in language-related data. Typically, these models are constructed on deep learning frameworks, like the transformer architecture \cite{b21}, which contributes to their remarkable performance across a spectrum of NLP assignments. The development of Bidirectional Encoder Representation Transformers (BERT) \cite{b9} and Generative Pre-trained Transformers (GPT) \cite{b10} has significantly propelled the NLP domain, ushering in the widespread adoption of pre-training and fine-tuning techniques \cite{b22}\cite{b23}. Extensive datasets are leveraged through task-specific aims and are commonly denoted as unsupervised training (although technically supervised, it lacks human-annotated labels) \cite{b24}\cite{b25}. Subsequently, it is standard procedure to fine-tune the pre-trained model to amplify the utility of the final model for downstream tasks. Noteworthy is the superior performance of these models over earlier deep learning approaches \cite{b26}, prompting a focus on the meticulous design of pre-training and fine-tuning procedures. Research endeavors have also shifted towards the ultimate objectives of machine learning, pre-training language models as intermediary components of self-directed learning tasks.

\subsection{Time Series Forecasting Tasks}
The various functions associated with time series can be categorized into multiple types, including prediction, anomaly identification, grouping, identification of change points, and segmentation. Prediction and anomaly detection are among the most commonly utilized applications. The task of time prediction involves examining the historical data patterns and trends in time series information to forecast future numerical changes using different time series prediction models. The forecasted future time point can be a singular value or a series of values over a specific time frame. This predictive task is frequently employed to analyze and predict time-dependent data such as stock prices, sales figures, temperature variations, and traffic volume. Given a sequence of values $Y =\{x_1, x_2,\ldots x_i, \ldots , x_N \}$, where the value corresponding to time stamp $t$ is $x_t$, the objective of prediction is to estimate $x_{t+1}$ based on $\{x_1, x_2, \ldots ,x_t\}$.

Traditional time series modeling, such as the ARIMA model, is commonly employed for forecasting time series data. The ARIMA model consists of three key components: the autoregressive model (AR), the differencing process, and the moving average model (MA). Essentially, the ARIMA model leverages historical data to make future predictions. The value of a variable at a specific time is influenced by its past values and previous random occurrences. This implies that the ARIMA model assumes the variable fluctuates around a general time trend, where the trend is shaped by historical values and the fluctuations are influenced by random events within a timeframe. Moreover, the general trend itself may not remain constant. In essence, the ARIMA model aims to uncover hidden patterns within the time series data using autocorrelation and differencing techniques. These patterns are then utilized for future predictions. ARIMA models are adept at capturing both trends and temporary, sudden, or noisy data, making them effective for a wide range of time series forecasting tasks.

If we temporarily set aside the distinction, the ARIMA model can be perceived as a straightforward fusion of the AR model and the MA model. In a formal manner, the equation for the ARIMA model can be represented as:
\begin{equation}
\begin{aligned}
x_t=& c+\varphi_1x_{t-1}+\varphi_2x_{t-2}+\ldots+\varphi_px_{t-p}+\\&\theta_1\epsilon_{t-1}+\theta_2\epsilon_{t-2}+\ldots+\theta_q\epsilon_{t-q}+\epsilon_t.
\end{aligned}
\end{equation}
In the equation provided, $x_t$ represents the time series data under examination, while $\varphi_1$ to $\varphi_p$ denote the coefficients of the autoregressive (AR) model that depict the connection between the present value and the value at the previous $p$ time instances. Similarly, $\theta_1$ to $\theta_q$ stand for the coefficients of the moving average (MA) model, which elucidate the relationship between the current value and the deviation at the last $q$ time points. The term $\epsilon_t$ symbolizes the deviation at time $t$, and $c$ is a constant term.

\subsection{LLMTIME}
A technique for predicting time series using Large Language Models (like GPT-3) involves converting numerical data into text and generating potential future trends through text predictions. Essentially, this approach converts time series data into a sequence of numbers and interprets time series prediction as forecasting the next token in a textual context, leveraging advanced pre-trained models and probabilistic functions like likelihood evaluation and sampling. The tokens play a crucial role in shaping patterns within tokenized sequences and determining the operations that the language model can comprehend.

Common methods for tokenization, such as Byte Pair Encoding (BPE), often break down a single number into segments that do not correspond directly to the number \cite{b27}. For instance, the GPT-3 tagger decomposes the number 42235630 into [422,35,630]. New language models like LLaMA \cite{b28} typically treat numbers as a single entity by default. LLMTIME adopts a method where it separates each digit of a number using spaces, tokenizes each digit individually, and uses commas to distinguish between different time steps in a time series. Since decimal points do not add value for a specific precision level, LLMTIME eliminates them during encoding to reduce the length of the context. For example, with a precision of two digits, we preprocess a time series by converting it from \[0.789,7.89,78.9,789.0\to\text{7 8 , 7 8 9 , 7 8 9 0 , 7 8 9 0 0}.\]

LLMTIME resizes the value to ensure that the rescaled time series value at the $\alpha$-percentile is 1. It refrains from scaling by the maximum value to allow the LLM to observe instances $(1-\alpha)$ where the quantity of numbers varies, and it mimics this pattern in its results to generate a larger value than it observes. Furthermore, LLMTIME endeavors to incorporate an offset $\beta$ computed from the percentile of the input data and fine-tunes these two parameters in accordance with the verification log-likelihood.

\subsection{Our Work}
Although LLMTIME may perform similarly to the traditional autoregressive model ARIMA on some datasets, its forecasting accuracy is lower than ARIMA when dealing with other datasets such as traditional and artificial signals that exhibit noisy almost periodic patterns. Moreover, as the data values increase over time, the forecasting performance of LLMTIME diminishes noticeably.

% \vspace{12pt}
\section{Experiments}
The zero-shot forecasting ability of LLMs is assessed by contrasting LLMTIME with well-known time series benchmarks across different baseline time series datasets. The model 'text-davinci-003' has been discontinued, further details can be found in \cite{b29}. Consequently, we exclusively employed the 'gpt-3.5-turbo-instruct' model for our assessments. In terms of deterministic metrics like MAE, LLMs exhibit inferior performance when compared to conventional forecasting techniques like ARIMA.

\begin{figure} [ht]
\centering
\subfloat[The effect of AirPassengersDataset based on LLMTIME. \label{a}]{
    \includegraphics[width=0.7\linewidth]{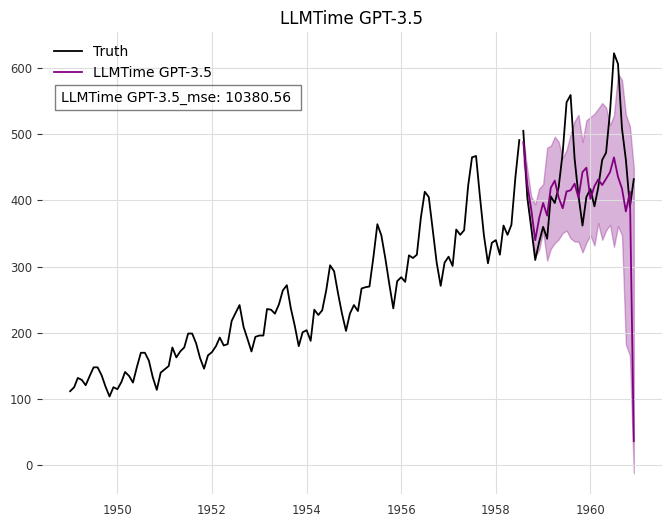}}
\\
\subfloat[The effect of AirPassengersDataset based on ARIMA. 
\label{b}]{
    \includegraphics[width=0.7\linewidth]{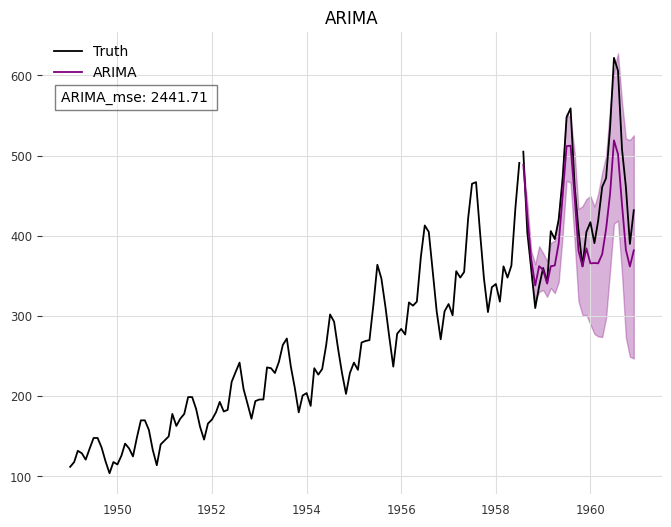} }
\caption{Figure~(\ref{a}) illustrates the impact of forecasting the AirPassengersDataset dataset using the LLMTIME model. It is evident that as the time series values increase gradually over time, the LLMTIME model predicts the final series value accurately but then sharply decreases, resulting in significantly poorer forecasting performance compared to the ARIMA model. This discrepancy is particularly noticeable in the FredMd dataset at Monash. Figure~(\ref{b}) displays the application of the ARIMA model for predicting the AirPassengersDataset dataset. Regardless of the waveform or periodicity of the predictions, ARIMA consistently produces excellent results. Moreover, ARIMA outperforms the LLMTIME model significantly in terms of overall Mean Squared Error (MSE) metrics. Further experimental results are elaborated in Appendix~\ref{A}.}
\label{fig1} 
\end{figure}

\subsection{Darts}\label{AA}
In Darts \cite{b30}, the darts.datasets module offers a variety of pre-existing time series datasets suitable for showcasing, testing, and experimentation purposes. These datasets typically consist of traditional time series data, with Darts specializing in time series forecasting and boasting a comprehensive forecasting system. We utilized LLMTIME and ARIMA models to conduct forecasting on eight datasets ('AirPassengersDataset', 'AusBeerDataset', 'GasRateCO2Dataset', 'MonthlyMilkDataset', 'SunspotsDataset', 'WineDataset', 'WoolyDataset', 'HeartRateDataset') integrated into Darts. A portion of the experimental outcomes is depicted in Fig.~\ref{fig1}. The AirPassengersDataset comprises classic airline data, documenting the total monthly count of international air passengers from 1949 to 1960, measured in thousands. All experimental findings are detailed in Appendix ~\ref{A}.

% \begin{figure} [ht]
% \centering
% \subfloat[The effect of AirPassengersDataset based on LLMTIME. \label{a}]{
%     \includegraphics[width=0.8\linewidth]{darts/AirPassengersDataset_LLMTIME.png}}
% \\
% \subfloat[The effect of AirPassengersDataset based on ARIMA. 
% \label{b}]{
%     \includegraphics[width=0.8\linewidth]{darts/AirPassengersDataset_ARIMA.png} }
% \caption{~(\ref{a}) shows the effect of predicting the AirPassengersDataset dataset based on the LLMTIME model, and we can notice that when the time series value gradually increases with the time step, the LLMTIME predicts the last series value and decreases sharply, and the actual forecasting effect is much worse than that of the ARIMA model. This phenomenon is more pronounced on FredMd dataset in Monash. Fig. 1 ~(\ref{b}) shows the use of the ARIMA model to predict the AirPassengersDataset dataset. Regardless of the predicted waveform or periodicity, ARIMA achieves perfect results. In addition, ARMIA is much better than the LLMTIME model in terms of overall MSE indicators.The rest of the experimental results are detailed in Appendix ~\ref{A}}
% \label{fig1} 
% \end{figure}

The experimental results demonstrate that the predictive performance of the LLMTIME approach utilizing GPT-3.5 on the Darts dataset is notably inferior to that of the ARIMA method. It is evident that as the data values increase steadily over time, the forecasting accuracy of LLMTIME diminishes significantly, a pattern observed in other data experiments as well.

\begin{figure} [ht]
\centering
\subfloat[The effect of FreMD dataset based on LLMTIME.\label{2a}]
{\includegraphics[width=0.7\linewidth]{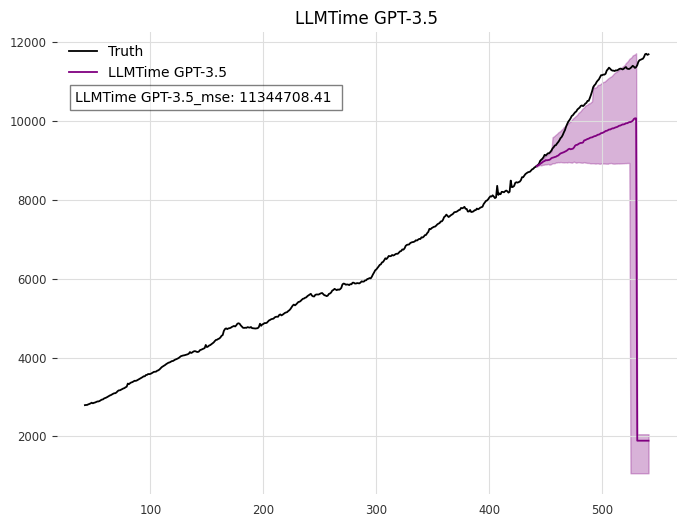}}
\\
\subfloat[The effect of CovidDeaths dataset based on LLMTIME.\label{2b}]{\includegraphics[width=0.7\linewidth]{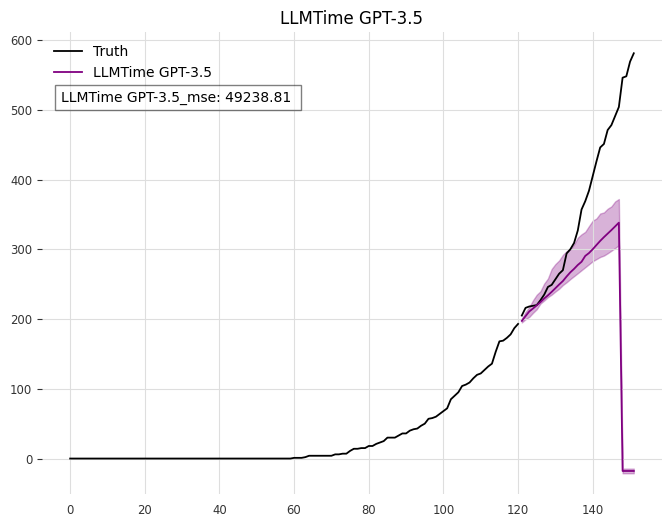}}
\caption{ When forecasting the final data point, it is evident from the provided graph that the performance based on LLMTIME experiences a significant decline.}
\label{fig2} 
\end{figure}

\subsection{Monash}\label{BB}
The Monash dataset, referenced as \cite{b31}, comprises 30 openly accessible datasets and baseline metrics for 12 predictive models. There are various versions of the datasets due to differences in frequency and handling of missing values, resulting in a total of 58 dataset variations. Additionally, the dataset collection includes a diverse range of real-world and competition time series datasets across different domains. Similar to the LLMTIME approach, we assessed GPT-3's zero-shot performance on the top 10 datasets with the most effective baseline model. A noticeable decline in the LLMTIME forecasting performance is evident in the FredMd dataset, which is a monthly database focused on Macroeconomic Research, as described in Working Paper 2015-012B by Michael W. McCracken and Serena Ng. This database aims to provide a comprehensive monthly macroeconomic dataset to facilitate empirical analysis that leverages "big data." (The outcomes are depicted in Fig.~\ref{fig2} for FreMD and CovidDeaths, with additional details in Appendix ~\ref{B}).

\begin{figure} [htbp]
\centering
\subfloat[Forecasts of total UK exports based on LLMTIME.\label{3a}]{
    \includegraphics[width=0.7\linewidth]{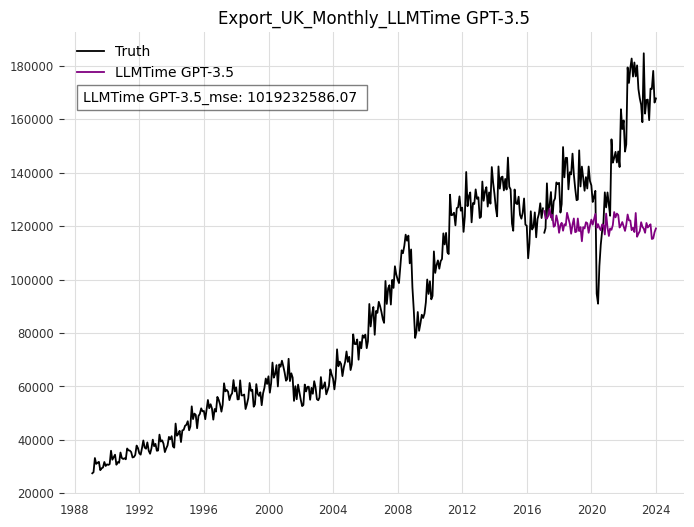}}
\\
\subfloat[Forecasts of total UK exports based on ARIMA.\label{3b}]{
    \includegraphics[width=0.7\linewidth]{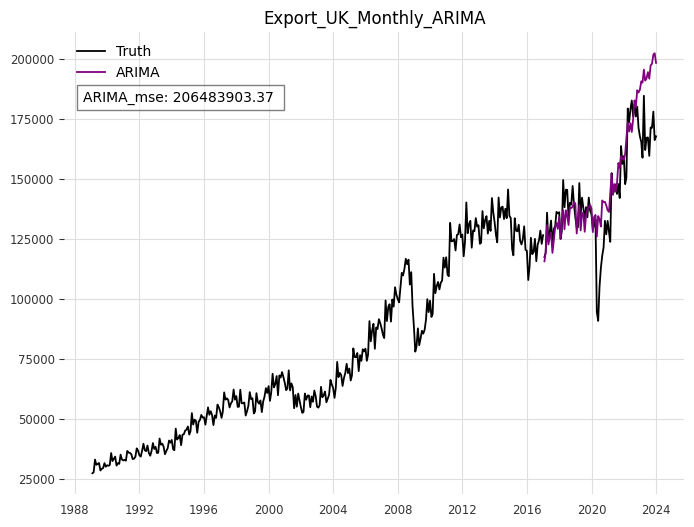}}
\caption{We predict the overall monetary worth of the UK's exports spanning from January 1989 to December 2023, quantified in millions of dollars and recorded on a monthly basis. (Refer to Appendix ~\ref{C} for more information).}
\label{fig3} 
\end{figure}

\begin{figure} [t!]
\centering
\subfloat[Predict the function $f(t) = cos(2\pi t)+cos(2t)+noise$ sequence values based on LLMTIME.\label{4a}]{
    \includegraphics[width=0.7\linewidth]{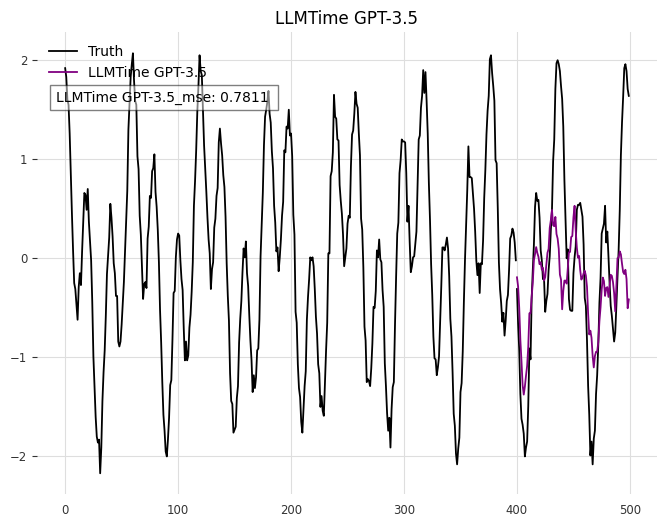}}
\\
\subfloat[Forecast the values of the noisy almost periodic function $f(t) = \cos(2\pi t) + \cos(2t) + \text{noise}$ using ARIMA.\label{4b}]{
    \includegraphics[width=0.7\linewidth]{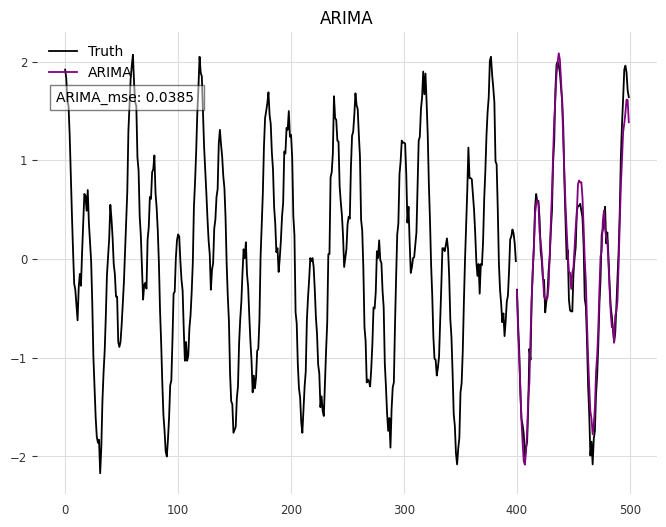}}
\caption{Utilizing the LLMTIME~(\ref{4a}) and ARIMA~(\ref{4b}) models, we anticipate the time series of the artificial signal $f(t) = cos(2\pi t)+cos(2t)+noise$, where $noise$ represents Gaussian noise with an average of 0 and a standard deviation of 0.1. Choose 500 data points from the function $f(t)$ within the range of 0 to 8$\pi$ to create a series, and forecast the values of the subsequent 100 points in the series based on the initial 400 data points.}
\label{fig4} 
\end{figure}

\subsection{Time series in economics}\label{CC}
In this section, we examined time series data related to the economy, specifically focusing on the total exports of six countries over the past few decades. The data was sourced from CEIE \cite{b36}, a comprehensive database offering economic information on over 213 countries and regions, encompassing macroeconomic indicators, industry-specific economics, and specialized data. Figure~\ref{fig3} illustrates the total export value of the United Kingdom from January 1989 to December 2023, denominated in millions of dollars and plotted against months.

\subsection{Data generated by noisy almost periodic functions}\label{DD}
A range of artificial signals were forecasted, and data was produced by adding noise to the almost periodic function \begin{equation}\label{eq:apf} f(t) = \cos(2\pi t) + \cos(2t) + noise,\end{equation} where the term $noise$ denotes Gaussian noise with a variance of $\sigma^2$. The almost periodic function has traditionally been significant in the development of harmonic analysis, leading to the formulation of Wiener's general harmonic analysis theory (GHA) \cite{b35}, and subsequently influencing the study of statistics of random processes.

The function \eqref{eq:apf} corresponds to an almost periodic function. In the field of mathematics, an almost periodic function is characterized by having multiple frequency components that do not share any common factor. Almost periodicity is a feature of dynamical systems that seem to revisit their trajectories in phase space, albeit not precisely. Several experiments were carried out by varying the standard deviations ($\sigma$). The outcomes of certain experiments ($\sigma$=0.1) are depicted in Fig.~\ref{fig4}.

\begin{figure}[h]
\centering
\includegraphics[width=0.8\linewidth]{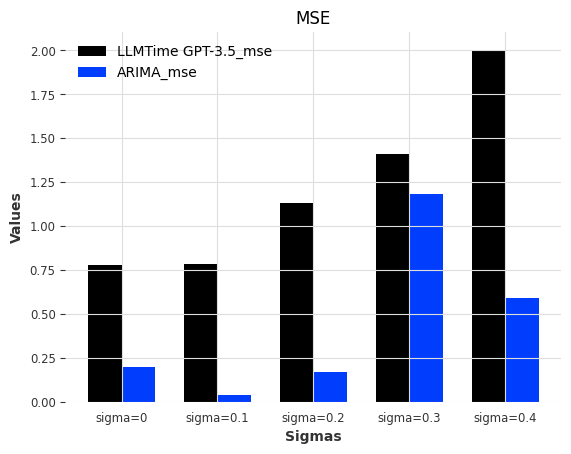}
\caption{The graph illustrates that the x-axis depicts the standard deviation of Gaussian noise, while the y-axis represents the MSE value. The black and blue bars correspond to LLMTIME and ARIMA, respectively. It can be inferred that across all four scenarios, the MSE associated with LLMTIME is significantly greater than that of ARIMA.}
\label{fig5}
\end{figure}

Furthermore, we conducted experiments on the four standard deviations of Gaussian noise using LLLTIME and ARIMA methods, and computed their mean square errors to generate Fig.~\ref{fig5}.

The comparison of the mean square errors indicates that the forecasts generated by the LLMTIME model generally exhibit higher errors compared to those produced by the ARMIR model. Additionally, the forecasting effectiveness graphs demonstrate that the ARMIR model outperforms the LLMTIME model significantly. For further details on the remaining experimental outcomes, please refer to Appendix ~\ref{D}.

\begin{figure} [t!]
\centering
\subfloat[Predict the function sequence values based on LLMTIME.\label{20a}]{
    \includegraphics[width=0.7\linewidth]{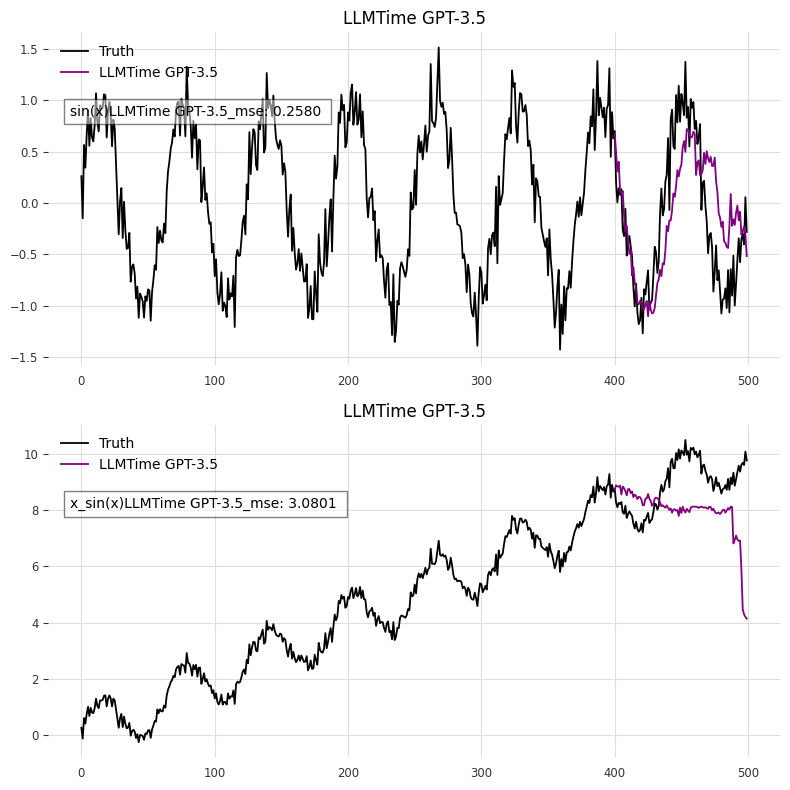}}
\\
\subfloat[Predict the function sequence values based on ARIMA.\label{20b}]{
    \includegraphics[width=0.7\linewidth]{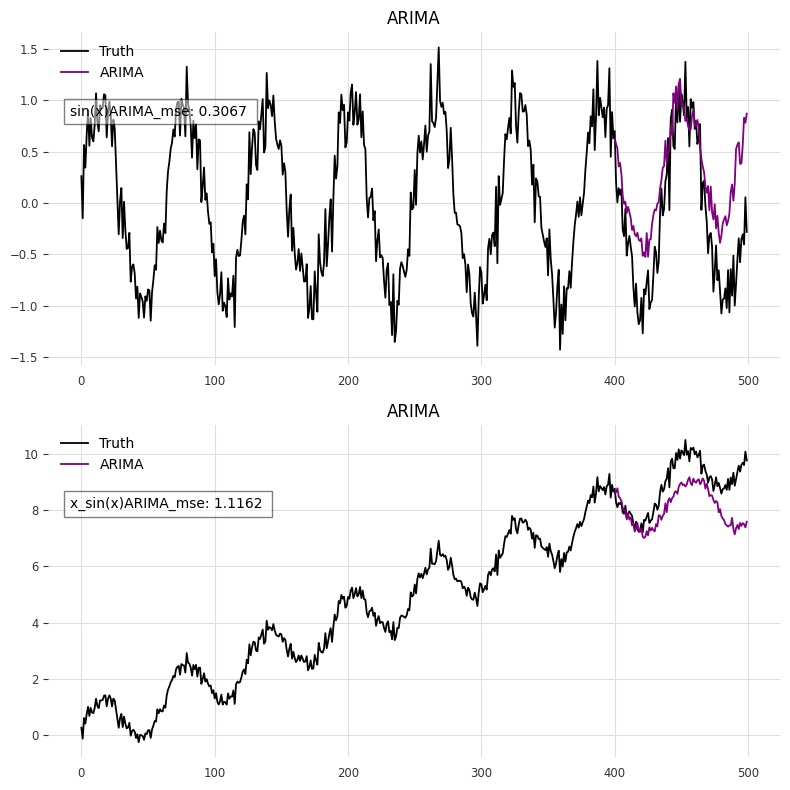}}
\caption{The upper part of the chart shows the genuine signal anticipated as $f(t) = sin(t) + noise$, while the lower part displays the genuine signal expected as $g(t) = sin(t) + 0.2t + noise$. The prediction performance of LLMTIME on $f(t)$ exhibits periodic characteristics, but it is evident that it fails to predict $g(t)$ accurately below. Here, the noise is Gaussian with a mean of zero and a standard deviation of 0.02.(See Appendix ~\ref{D} for details.)}
\label{fig20} 
\end{figure}

Furthermore, we examined the sine wave and the combination of sinusoidal and linear functions. The findings indicate that LLMTIME is effective only for time series that exhibit complete periodicity, while it struggles to forecast non-periodic signals. Refer to Fig.~\ref{fig20} for more information.

\begin{table}[htbp]
\caption{The MSE in some datasets based on LLMTIME and ARIMA.}
\begin{center}
\begin{tabular}{p{2.5cm}<{\centering}p{2.5cm}<{\centering}p{2cm}<{\centering}}
% \begin{tabular}{c|c|c}

\multicolumn{3}{c}{\textbf{DATASETS \hspace{0.5cm} LLMTIME\_MAE \hspace{0.5cm} ARIMA\_MSE}} \\
\hline
\hline

AirPassengers& 10380.56&\textbf{2441.71} \\
 AusBeer&\textbf{454.15}  &754.21\\
 GasRateCO2& 32.71 &\textbf{8.25}\\
 MonthlyMilk& 6023.25 &\textbf{1670.71}\\
 Sunspots& 1485.97 &\textbf{1007.62}\\
 Wine& 31683665.37 &\textbf{8826996.13}\\
 Wooly&628419.65  &\textbf{587217.59}\\
 HeartRate&87.95  &\textbf{53.22}\\
\hline

 Cif2016 & 98009.38 & \textbf{3173.05}\\
 CovidDeaths& 492338.81 & \textbf{10747.88}\\
 ElectricityDemand& 983958.23 & \textbf{720077.08} \\
 FredMd& 11344708.41 & \textbf{574689.74}\\
 Hospital& \textbf{18.79} & 22.57\\
 Nn5Weekly& \textbf{1964.31} & 3920.95\\
 PedestrianCounts& \textbf{306676.68} & 1082241.58\\
%  Saugeenday& \textbf{9.03} & 413.76\\
 TourismMonthly& 3033529.57 & \textbf{346055.66}\\
 TrafficWeekly&\textbf{1.28 }&1.31\\
 UsBirths& 1021661.84 &\textbf{411167.05}\\
\hline

\textit{$\sigma$=0$^{\mathrm{*}}$}& 0.7762&\textbf{0.1997}\\
 \textit{$\sigma$=0.1} & 0.7811 &\textbf{0.0385}\\
 \textit{$\sigma$=0.2} & 1.1289 &\textbf{0.1708}\\
 \textit{$\sigma$=0.3} & 1.4096 &\textbf{1.1824}\\
 \textit{$\sigma$=0.4} & 2.0037 &\textbf{0.5909}\\
\hline

\multicolumn{3}{l}{$^{\mathrm{*}}$The synthetic signal is $f(t) = cos(2\pi t)+cos(2t)+noise$,} \\
\multicolumn{3}{l}{where $\sigma$ denotes the standard deviation of Gaussian noise.}
\end{tabular}
\label{tab1}
\end{center}
\end{table}

Experimental outcomes obtained from the aforementioned datasets allow us to calculate the Mean Squared Error (MSE) between the predicted outcomes and the actual data. Our analysis indicates that, in most instances, the predictive accuracy of the LLMTIME model is inferior to that of the ARIMA model. Refer to Table \ref{tab1} for details.

\section{Conclusion}
% We have demonstrated that the LLMTIME model can only predict data with obvious periodic characteristics.For data that are superimposed by trend and cyclical components, it does not produce excellent forecasting and is not as good as the traditional time series model ARIMA. After the period component is superimposed with the trend component, LLMTIME does not accurately capture this information. In the scenario where there is a mixture of periodicity and this trend, the periodic information of the signal buries the trend information. In other words, LLMTIME does not have the ability to analyze such a delicate periodic structure, or a slightly more complex periodicity. In addition, when the trend of real data is monotonically rising or falling, the forecasting effect of LLMTIME will decline sharply when predicting the last few time steps. The specific causes of this phenomenon need to be further explored. In general, the use of LLMs as pre-trained time series forecasting models has certain challenges and difficulties, which is also a promising direction for future research. We hope there will be better ways to link LLMs research to time series forecasting.

It has been observed that the LLMTIME model encounters difficulties in accurately predicting datasets that contain both trend and cyclical elements, as well as in cases where the signal consists of intricate frequency components. In comparison to the conventional ARIMA time series model, the LLMTIME model is less dependable. Despite the challenges and limitations associated with using LLMs as pre-trained models for time series prediction, there is potential for further investigation in this area. We are eager to explore improved methods for integrating LLMs research into time series forecasting.

{\bf{Acknowledgment}}: We thank the referees of this paper for their very helpful comments. The code of this paper is available at: \href{https://github.com/crSEU/llmtime_VS_arima}{https://github.com/crSEU/llmtime\_VS\_arima}

%This paper is based on the LLMTIME model for further research and exploration. And this work was strongly supported by the Data Science and Data Engineering Laboratory of Southeast University.

% \newpage

\clearpage
\section{Appendix}
% Darts Dataset
\subsection{Detailed experimental results of Darts dataset}\label{A}
\vspace{36pt}
\begin{figure} [ht]
\centering
\subfloat[AirPassengersDataset(LLMTIME)]{
    \includegraphics[width=0.5\linewidth]{darts/AirPassengersDataset_LLMTIME.png}}
\subfloat[AirPassengersDataset(ARIMA)]{
    \includegraphics[width=0.5\linewidth]{darts/AirPassengersDataset_ARIMA.png}}\\
\subfloat[AusBeerDataset (LLMTIME)]{
    \includegraphics[width=0.5\linewidth]{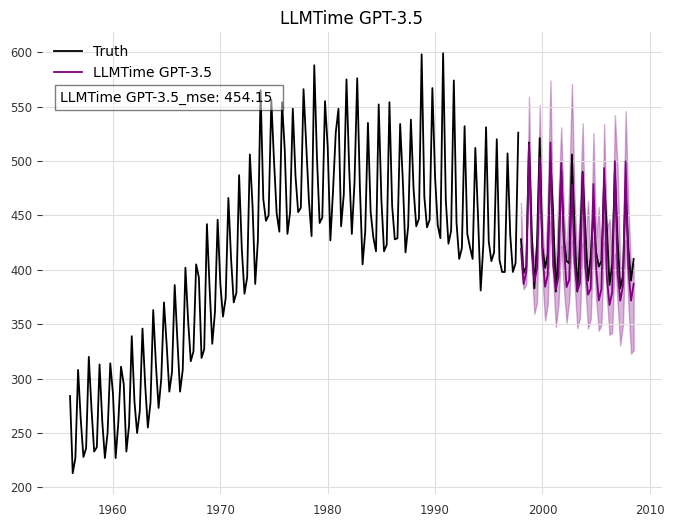}}
\subfloat[AusBeerDataset (ARIMA)]{
    \includegraphics[width=0.5\linewidth]{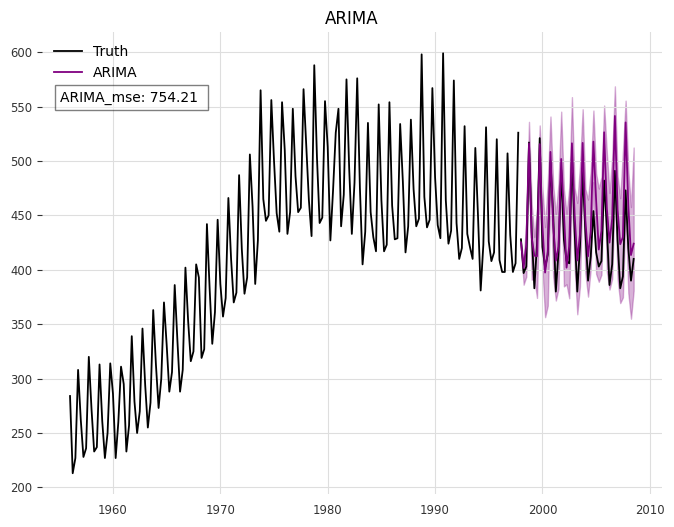}}\\
\subfloat[GasRateCO2Dataset (LLMTIME)]{
    \includegraphics[width=0.5\linewidth]{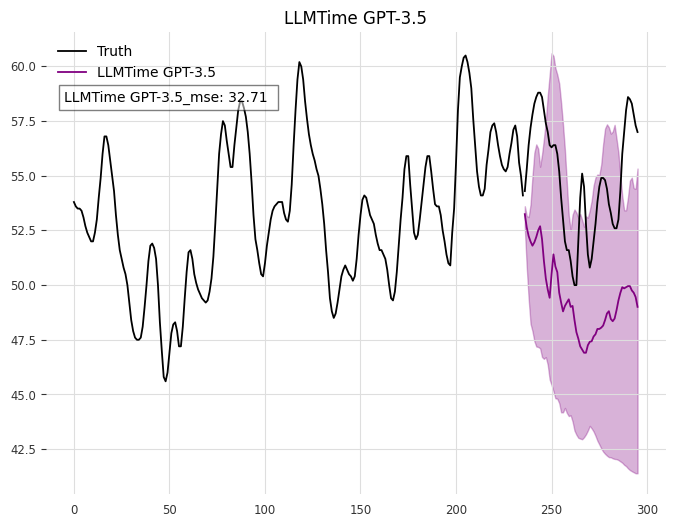}}
\subfloat[GasRateCO2Dataset (ARIMA)]{
    \includegraphics[width=0.5\linewidth]{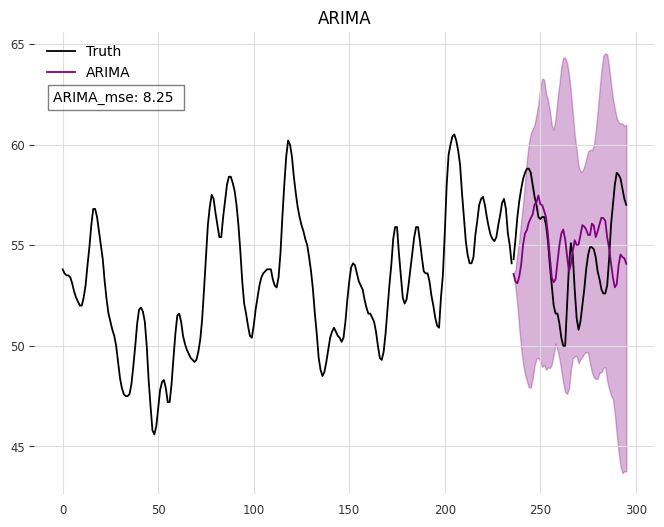}}\\
\subfloat[HeartRateDataset (LLMTIME)]{
    \includegraphics[width=0.5\linewidth]{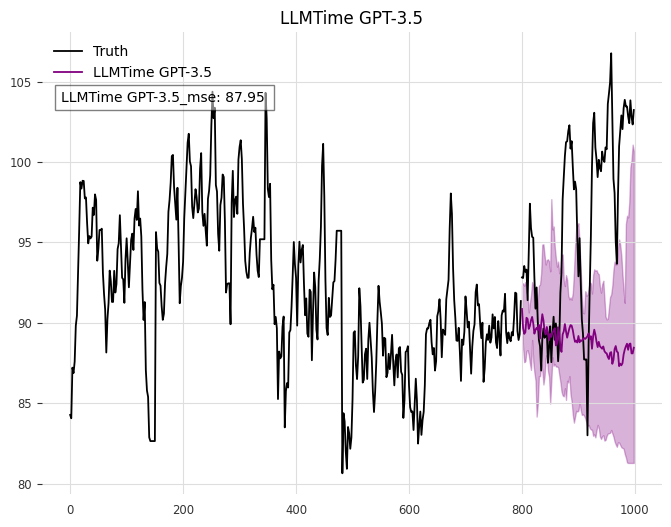}}
\subfloat[HeartRateDataset (ARIMA)]{
    \includegraphics[width=0.5\linewidth]{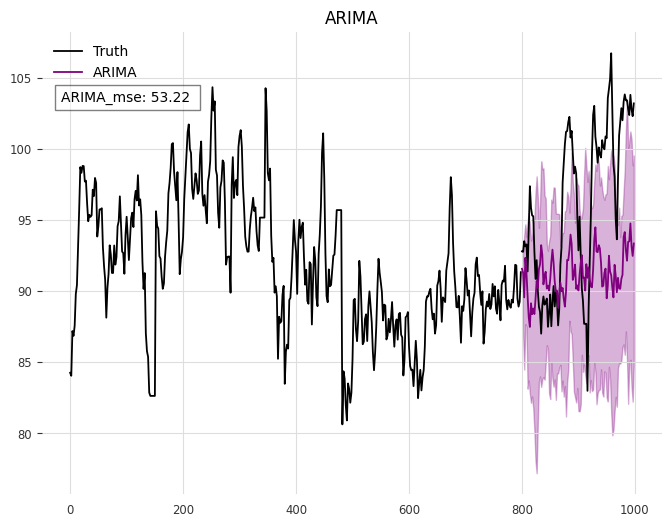}}
\caption{Experimental results of 4 datasets in Darts.}
\label{fig6} 
\end{figure}

\begin{figure} [t]
\centering
\subfloat[MonthlyMilkDataset (LLMTIME)]{
    \includegraphics[width=0.5\linewidth]{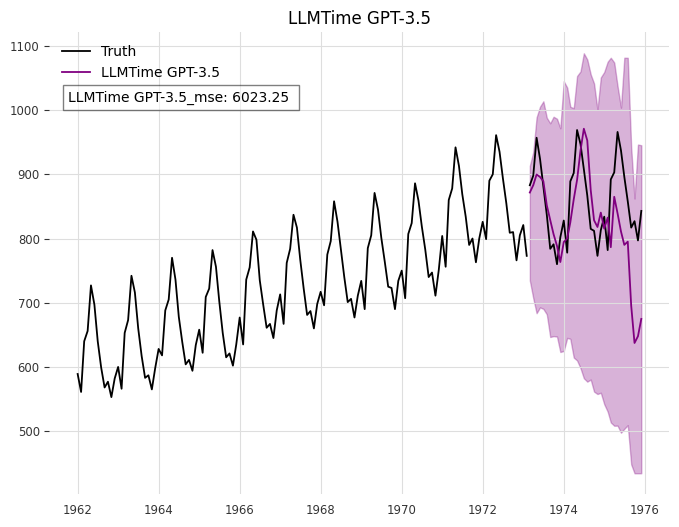}}
\subfloat[MonthlyMilkDataset (ARIMA)]{
    \includegraphics[width=0.5\linewidth]{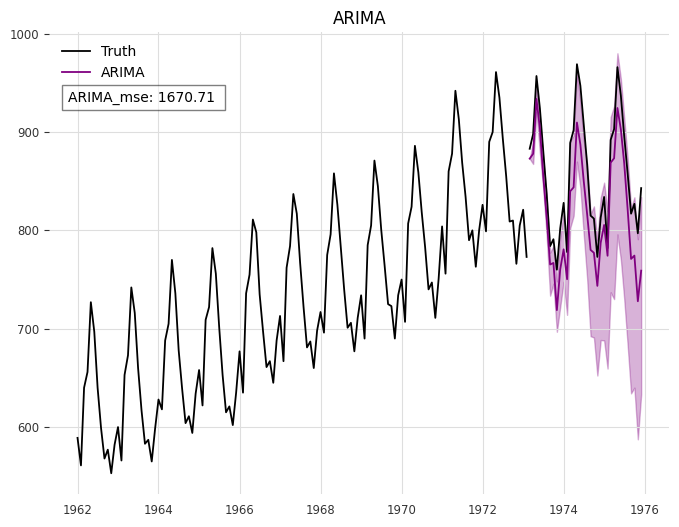}}\\
\subfloat[SunspotsDataset (LLMTIME)]{
    \includegraphics[width=0.5\linewidth]{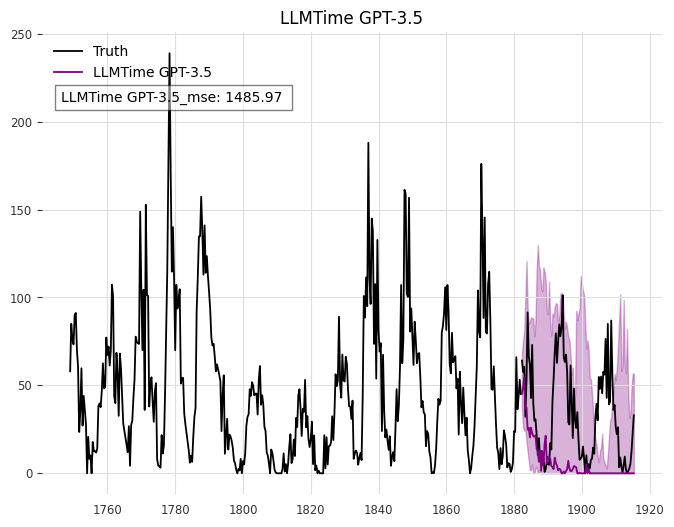}}
\subfloat[SunspotsDataset (ARIMA)]{
    \includegraphics[width=0.5\linewidth]{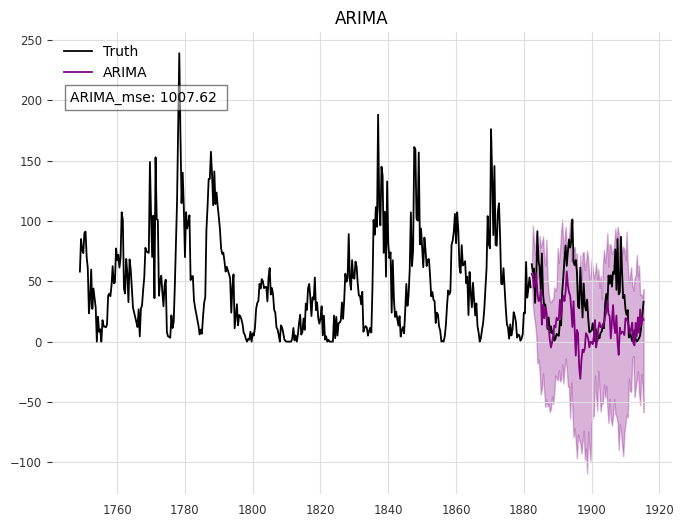}}\\
\subfloat[WineDataset (LLMTIME)]{
    \includegraphics[width=0.5\linewidth]{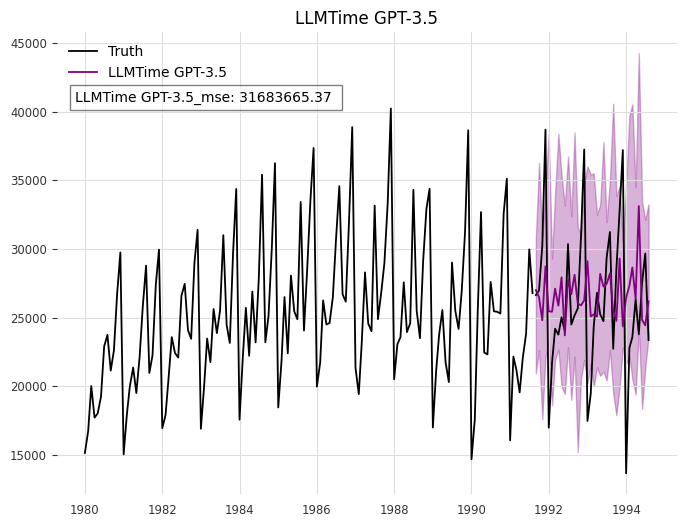}}
\subfloat[WineDataset (ARIMA)]{
    \includegraphics[width=0.5\linewidth]{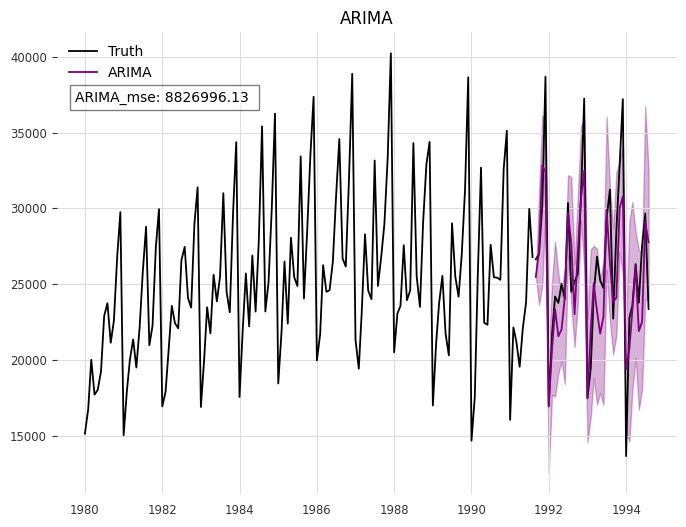}}\\
\subfloat[WoolyDataset (LLMTIME)]{
    \includegraphics[width=0.5\linewidth]{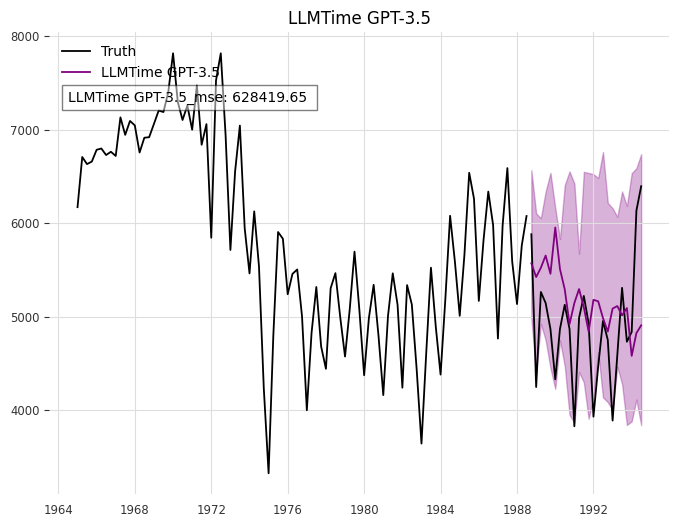}}
\subfloat[WoolyDataset (ARIMA)]{
    \includegraphics[width=0.5\linewidth]{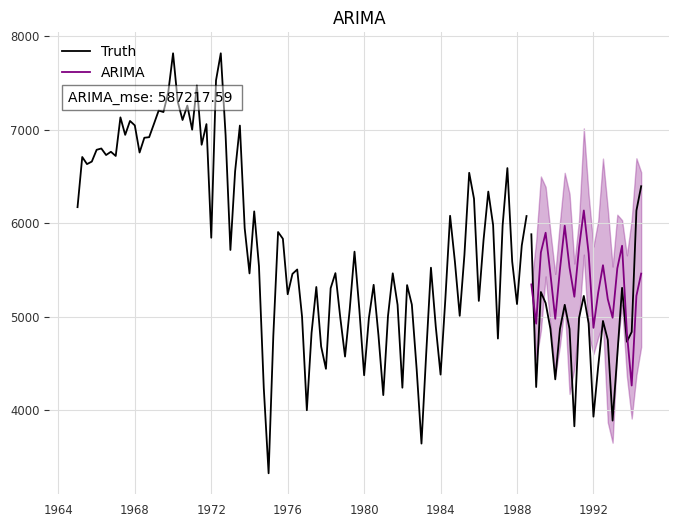}}

\caption{Experimental results of 4 datasets in Darts.}
\label{fig7} 
\end{figure}

\clearpage
% Monash Dataset
\subsection{Detailed experimental results of Monash dataset}\label{B}

\vspace{20pt}
\begin{figure} [ht]
\centering
\subfloat[Cif2016 (LLMTIME)]{
    \includegraphics[width=0.45\linewidth]{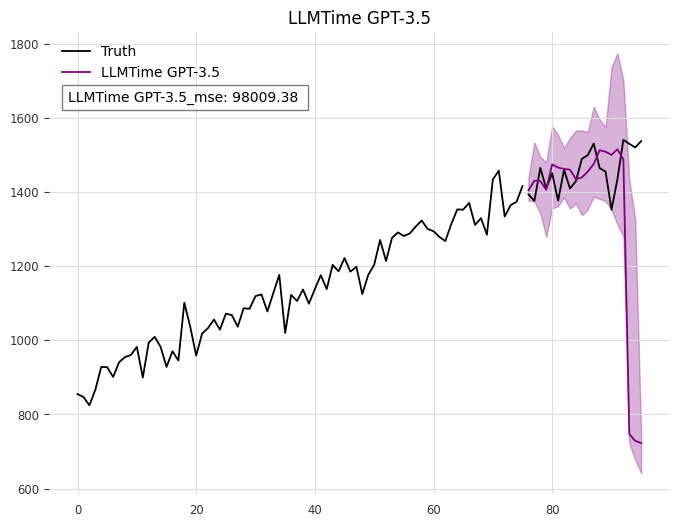}}
\subfloat[Cif2016 (ARIMA)]{
    \includegraphics[width=0.45\linewidth]{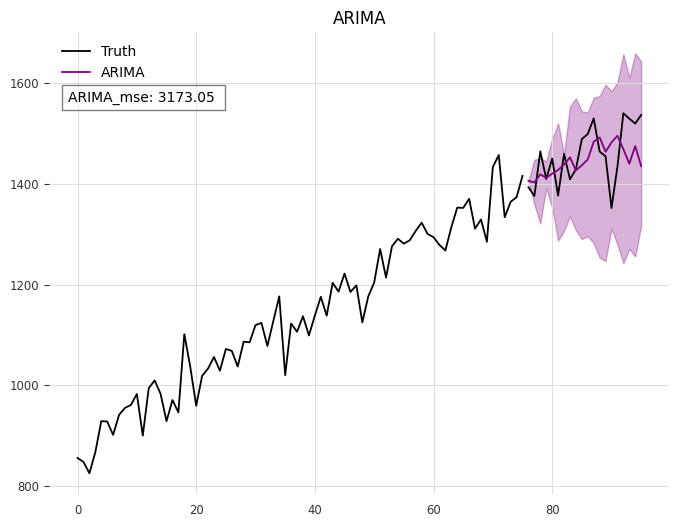}}\\
\subfloat[CovidDeaths (LLMTIME)]{
    \includegraphics[width=0.45\linewidth]{monash/covid_deaths_llmtime.png}}
\subfloat[CovidDeaths (ARIMA)]{
    \includegraphics[width=0.45\linewidth]{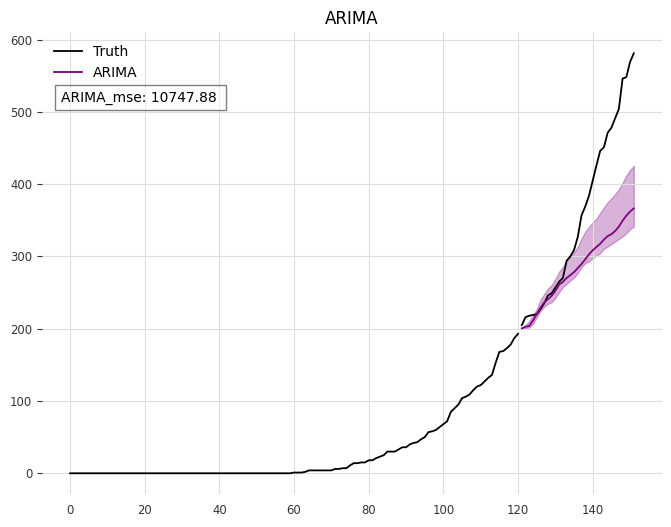}}\\
\subfloat[ElectricityDemand (LLMTIME)]{
    \includegraphics[width=0.45\linewidth]{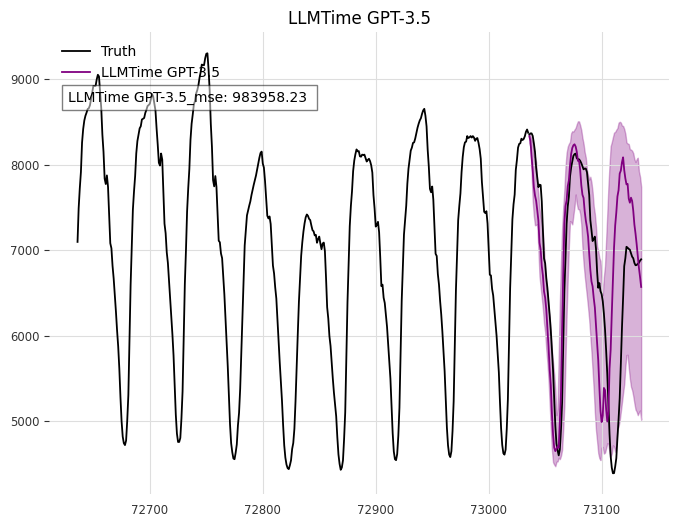}}
\subfloat[ElectricityDemand (ARIMA)]{
    \includegraphics[width=0.45\linewidth]{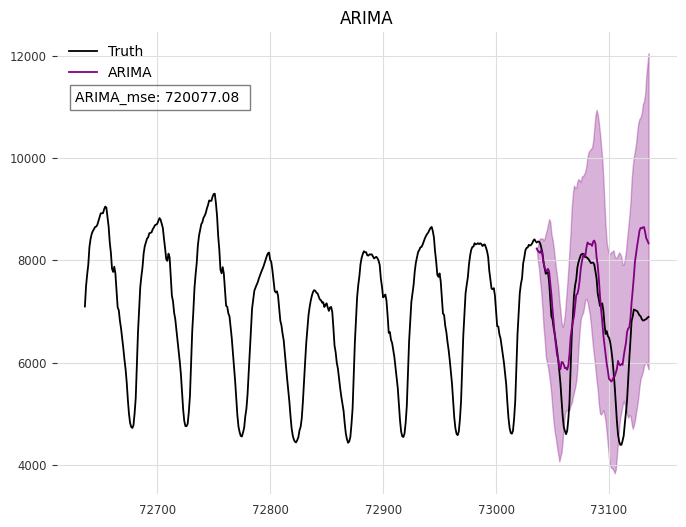}}\\
\subfloat[FredMd (LLMTIME)]{
    \includegraphics[width=0.45\linewidth]{monash/fred_md_llmtime.png}}
\subfloat[FredMd (ARIMA)]{
    \includegraphics[width=0.45\linewidth]{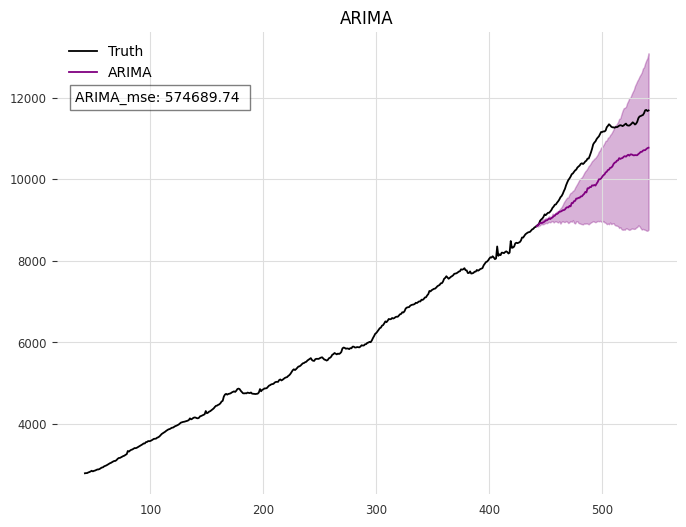}}\\
\subfloat[Hospital (LLMTIME)]{
    \includegraphics[width=0.45\linewidth]{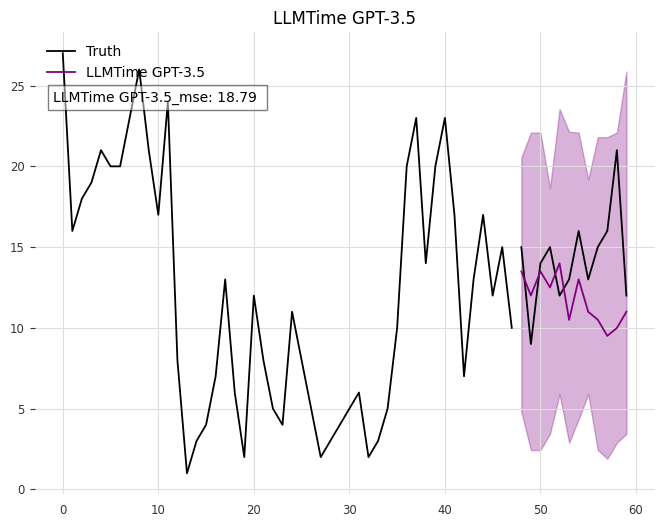}}
\subfloat[Hospital (ARIMA)]{
    \includegraphics[width=0.45\linewidth]{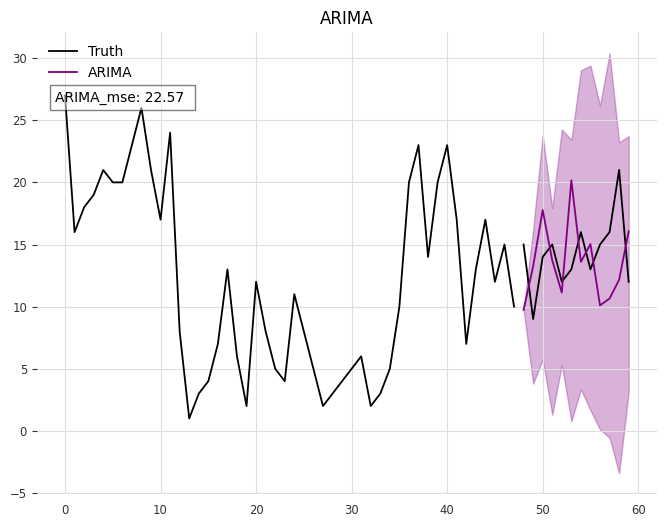}}\\
\caption{Experimental results of 5 datasets in Monash.}
\label{fig8} 
\end{figure}

\begin{figure} [htbp]
\centering

\subfloat[Nn5Weekly (LLMTIME)]{
    \includegraphics[width=0.45\linewidth]{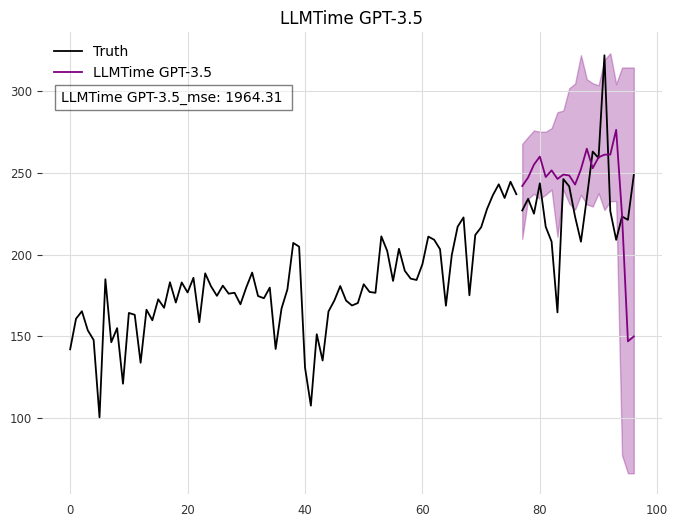}}
\subfloat[Nn5Weekly (ARIMA)]{
    \includegraphics[width=0.45\linewidth]{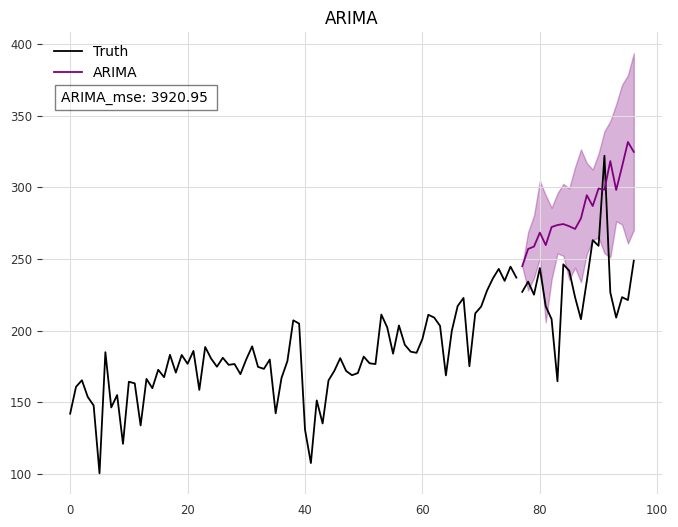}}\\
\subfloat[PedestrianCounts (LLMTIME)]{
    \includegraphics[width=0.45\linewidth]{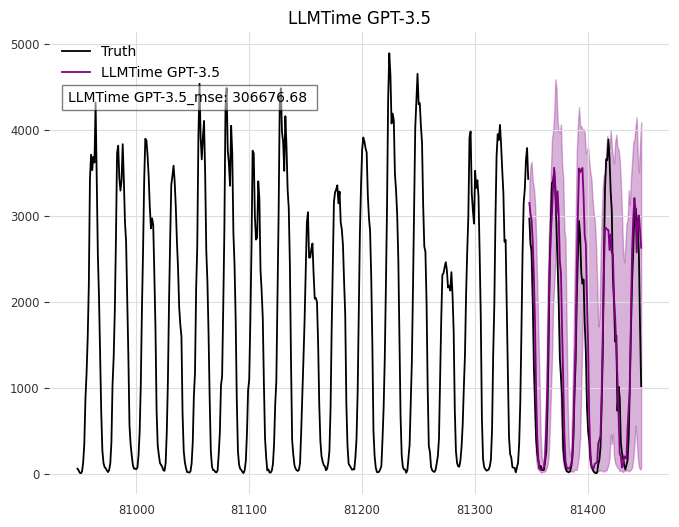}}
\subfloat[PedestrianCounts (ARIMA)]{
    \includegraphics[width=0.45\linewidth]{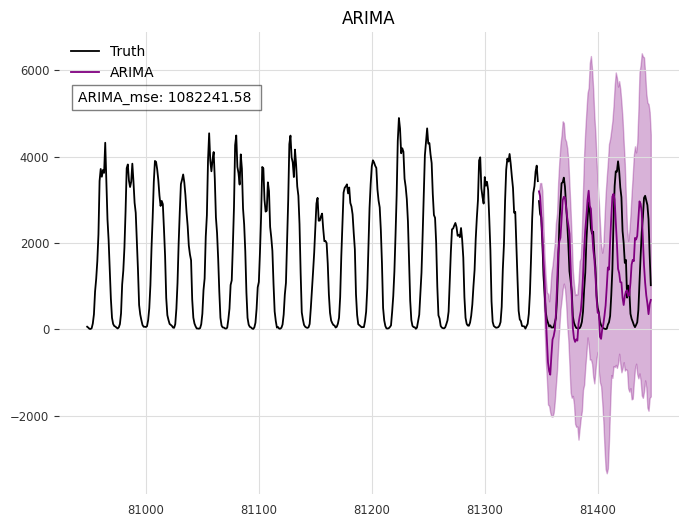}}\\
% \subfloat[Saugeenday(LLMTIME)]{
%     \includegraphics[width=0.5\linewidth]{monash/saugeenday_llmtime.png}}
% \subfloat[Saugeenday(ARIMA)]{
%     \includegraphics[width=0.5\linewidth]{monash/saugeenday_arima.png}}\\
\subfloat[TourismMonthly (LLMTIME)]{
    \includegraphics[width=0.45\linewidth]{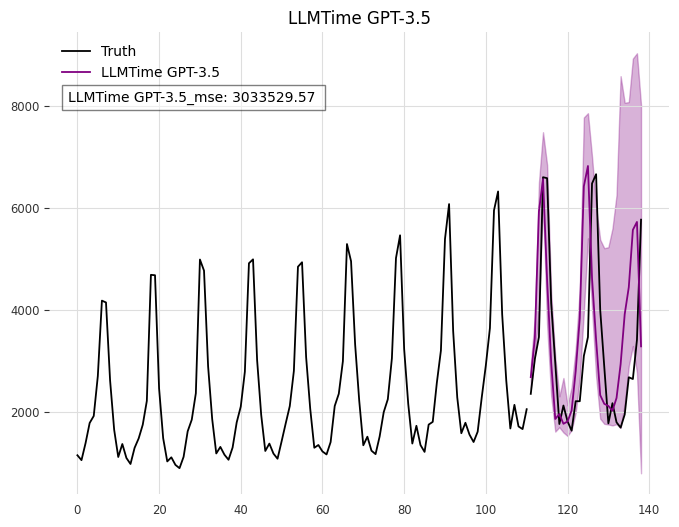}}
\subfloat[TourismMonthly (ARIMA)]{
    \includegraphics[width=0.45\linewidth]{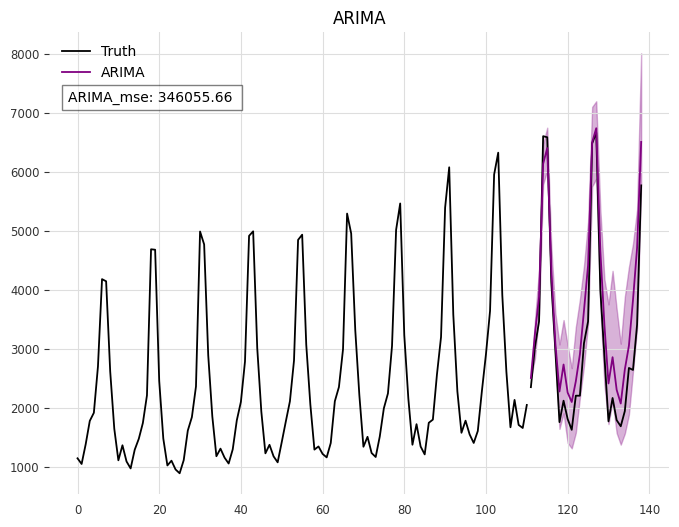}}\\
\subfloat[TrafficWeekly (LLMTIME)]{
    \includegraphics[width=0.45\linewidth]{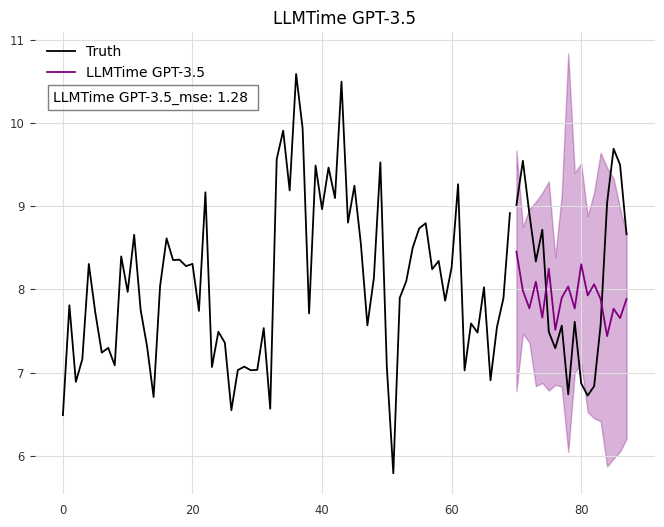}}
\subfloat[TrafficWeekly (ARIMA)]{
    \includegraphics[width=0.45\linewidth]{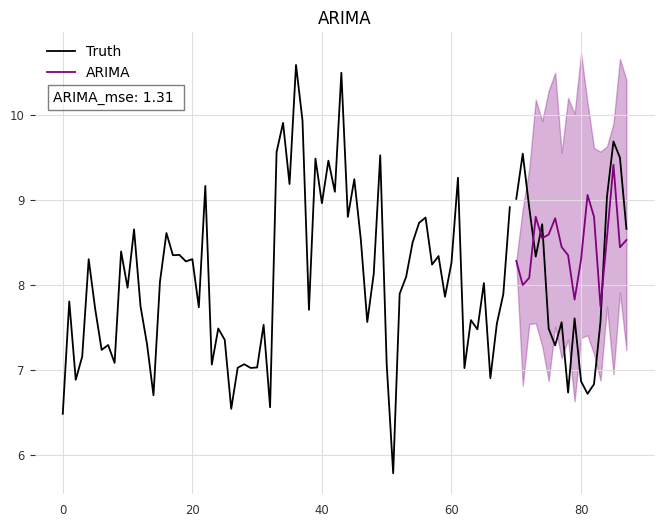}}\\
\subfloat[UsBirths (LLMTIME)]{
    \includegraphics[width=0.45\linewidth]{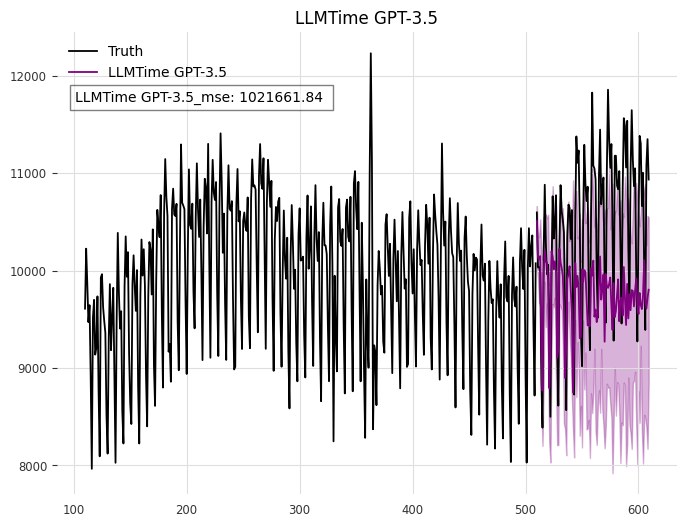}}
\subfloat[UsBirths (ARIMA)]{
    \includegraphics[width=0.45\linewidth]{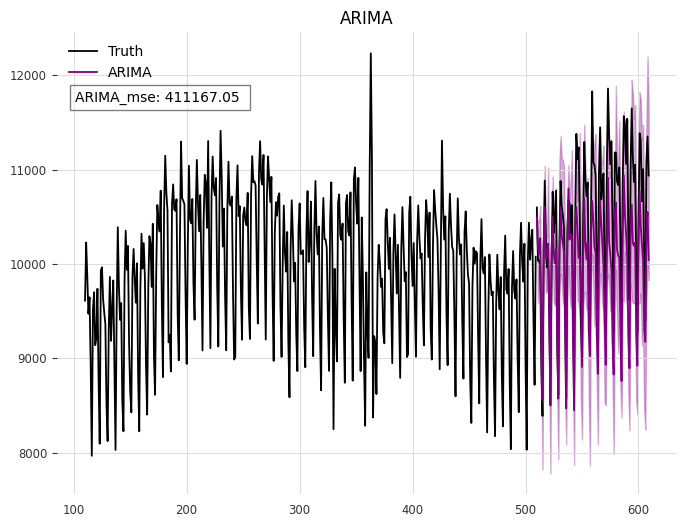}}\\
\caption{Experimental results of 5 datasets in Monash.}
\label{fig9} 
\end{figure}
\clearpage
\subsection{Detailed experimental results of time series in economics}\label{C}
\vspace{10pt}
\begin{figure} [htbp]
\centering
\subfloat[American exports (LLMTIME)]{
    \includegraphics[width=0.45\linewidth]{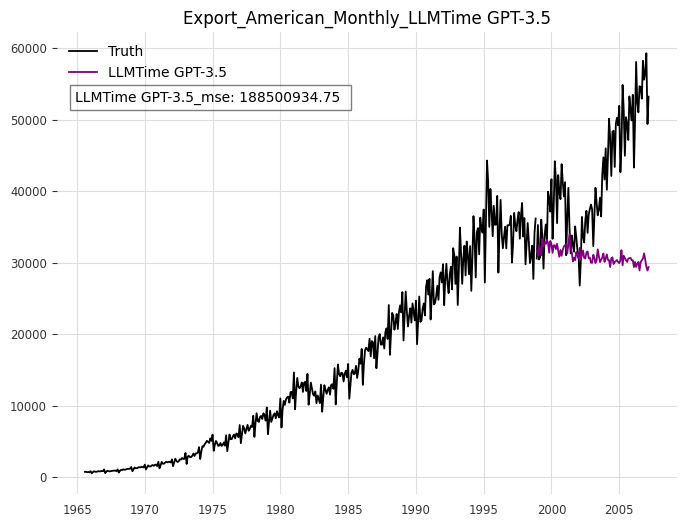}}
\subfloat[American exports (ARIMA)]{
    \includegraphics[width=0.45\linewidth]{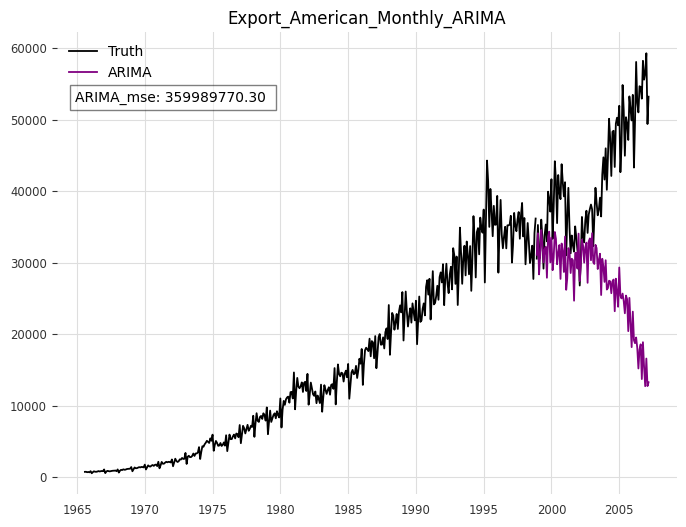}}\\
\subfloat[China exports (LLMTIME)]{
    \includegraphics[width=0.45\linewidth]{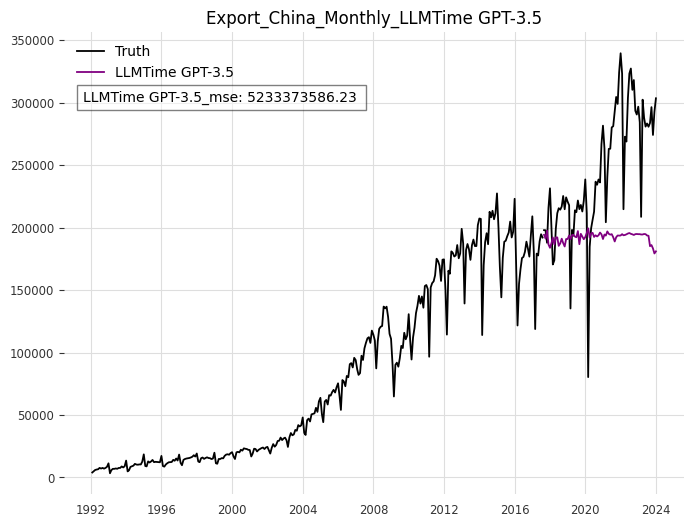}}
\subfloat[China exports (ARIMA)]{
    \includegraphics[width=0.45\linewidth]{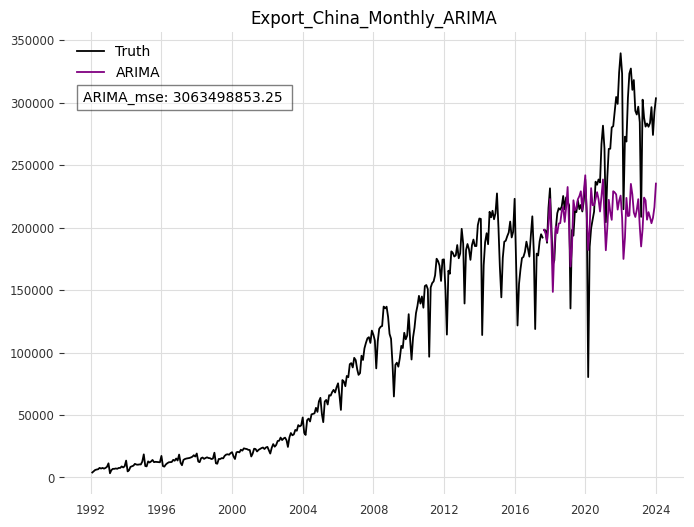}}\\
\subfloat[France exports (LLMTIME)]{
    \includegraphics[width=0.45\linewidth]{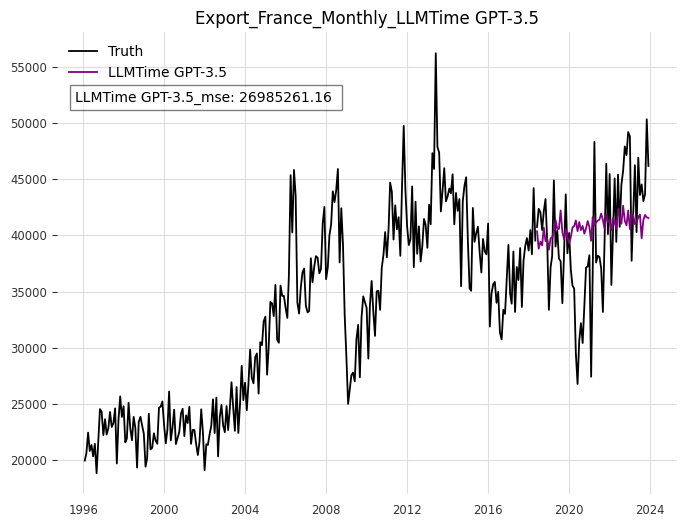}}
\subfloat[France exports (ARIMA)]{
    \includegraphics[width=0.45\linewidth]{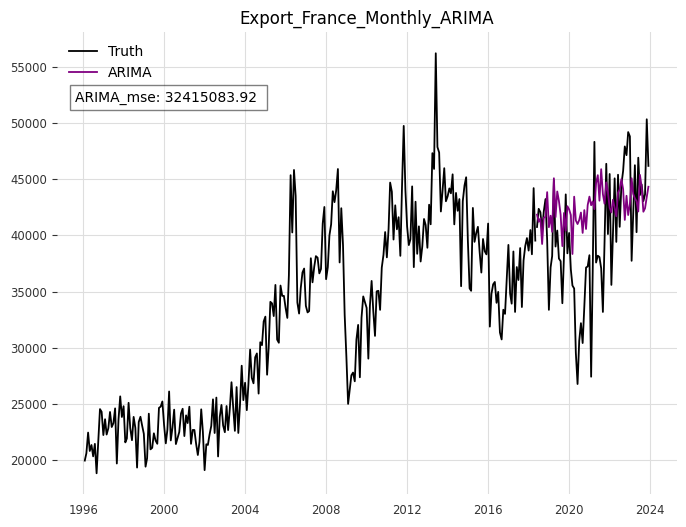}}\\
\subfloat[HongKong exports (LLMTIME)]{
    \includegraphics[width=0.45\linewidth]{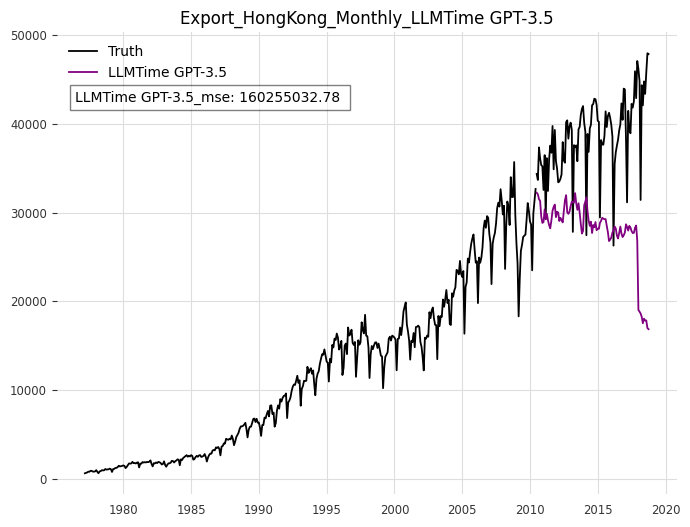}}
\subfloat[HongKong exports (ARIMA)]{
    \includegraphics[width=0.45\linewidth]{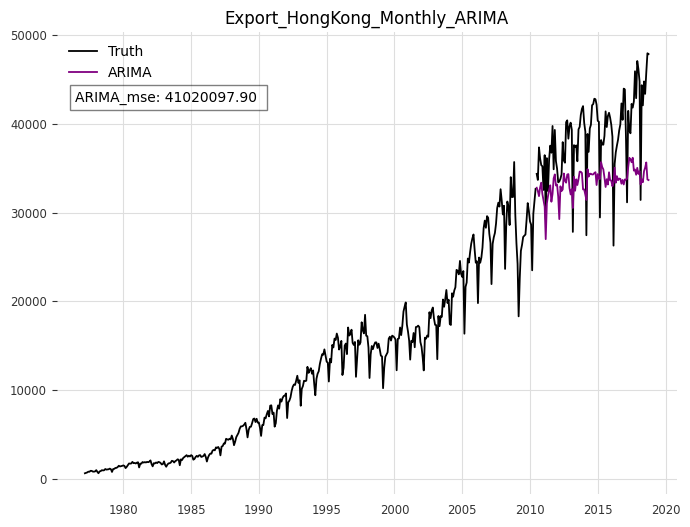}}\\
\subfloat[Japan exports (LLMTIME)]{
    \includegraphics[width=0.45\linewidth]{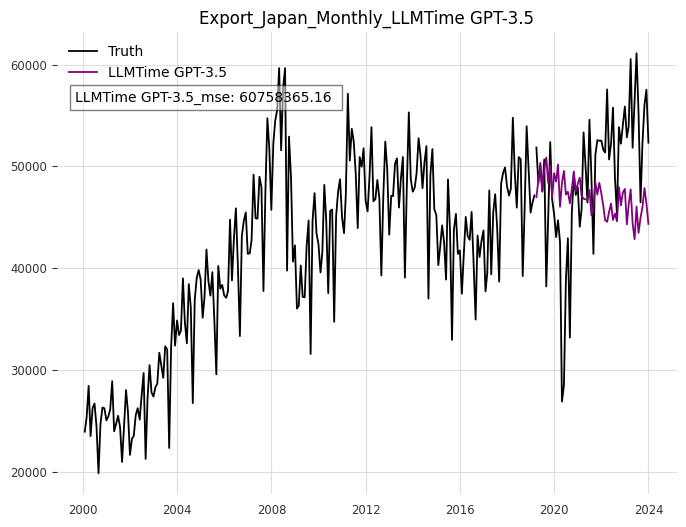}}
\subfloat[Japan exports (ARIMA)]{
    \includegraphics[width=0.45\linewidth]{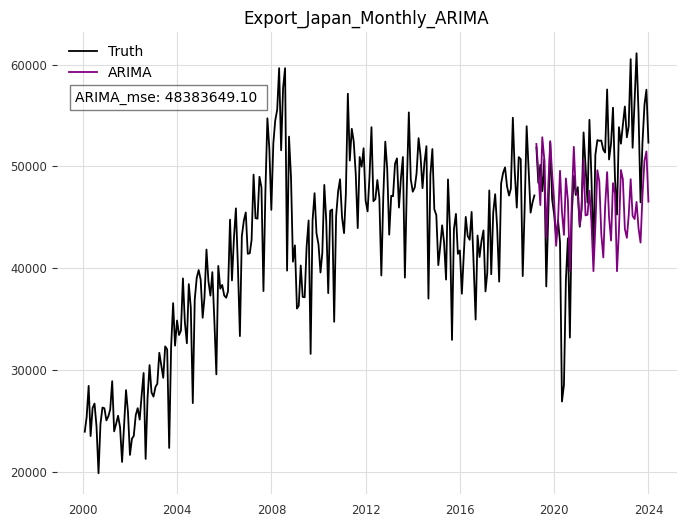}}\\

\caption{Experimental results of economics dataset.}
\label{fig10} 
\end{figure}
\vspace{45pt}
% Synthetic Dataset
\clearpage
\subsection{Detailed experimental results of Synthetic dataset}\label{D}
\vspace{10pt}
\begin{figure} [ht]
\centering
\subfloat[$\sigma=0$ (LLMTIME)]{
    \includegraphics[width=0.45\linewidth]{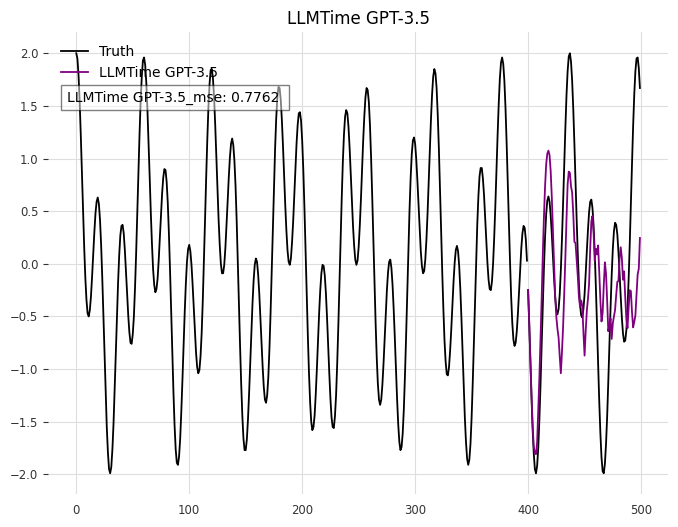}}
\subfloat[$\sigma=0$ (ARIMA)]{
    \includegraphics[width=0.45\linewidth]{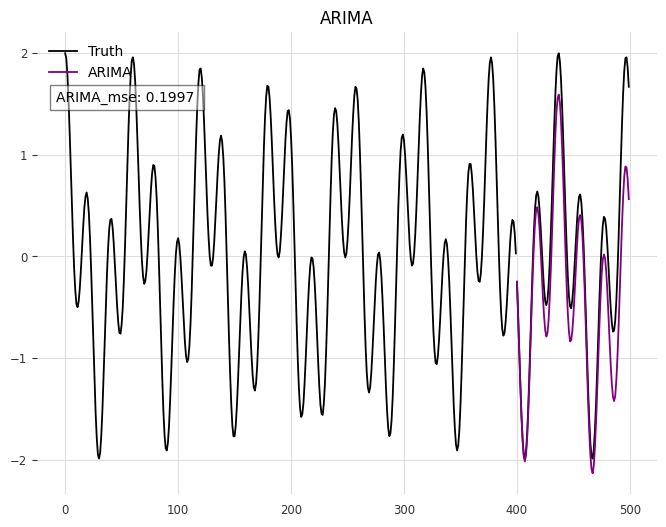}}\\
\subfloat[$\sigma=0.1$ (LLMTIME)]{
    \includegraphics[width=0.45\linewidth]{cosines/cosines_llmtime_1.png}}
\subfloat[$\sigma=0.1$ (ARIMA)]{
    \includegraphics[width=0.45\linewidth]{cosines/cosines_arima_1.png}}\\
\subfloat[$\sigma=0.2$ (LLMTIME)]{
    \includegraphics[width=0.45\linewidth]{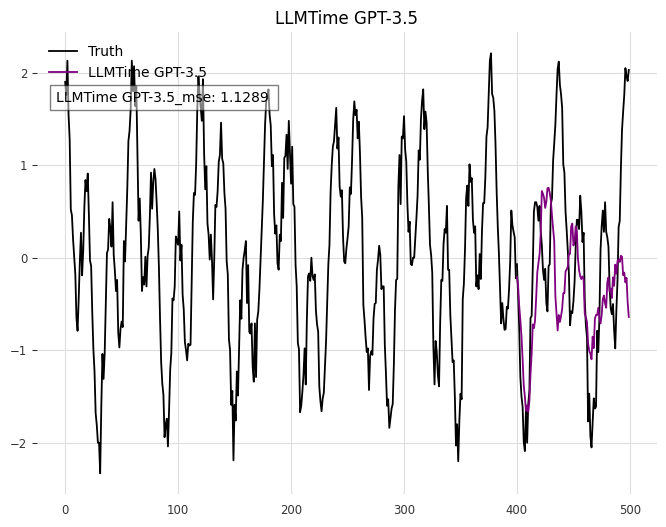}}
\subfloat[$\sigma=0.2$ (ARIMA)]{
    \includegraphics[width=0.45\linewidth]{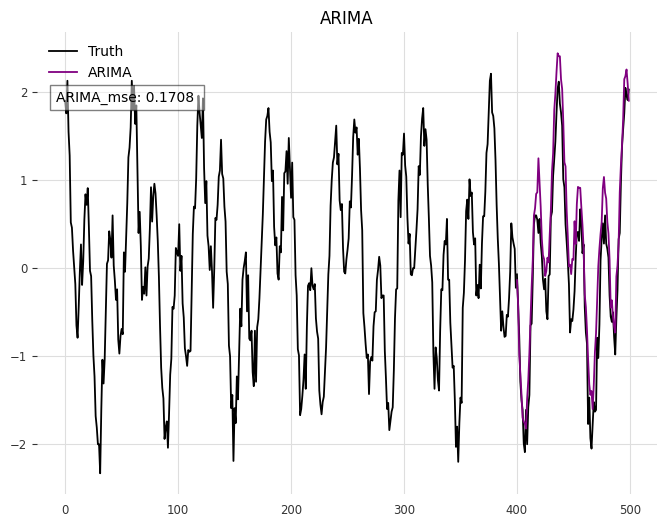}}\\
\subfloat[$\sigma=0.3$ (LLMTIME)]{
    \includegraphics[width=0.45\linewidth]{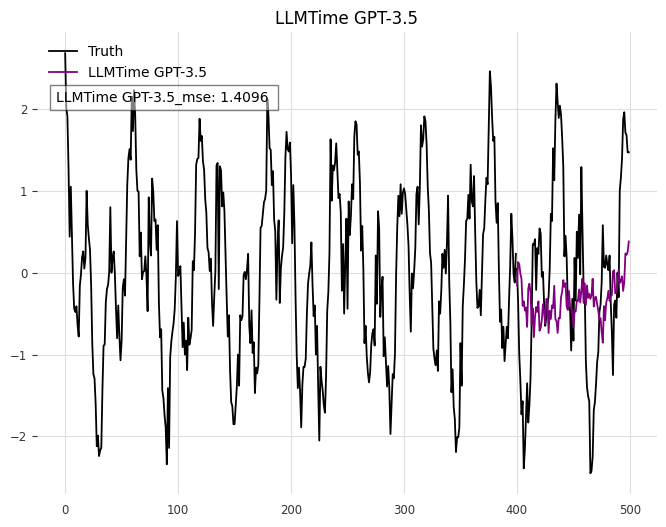}}
\subfloat[$\sigma=0.3$ (ARIMA)]{
    \includegraphics[width=0.45\linewidth]{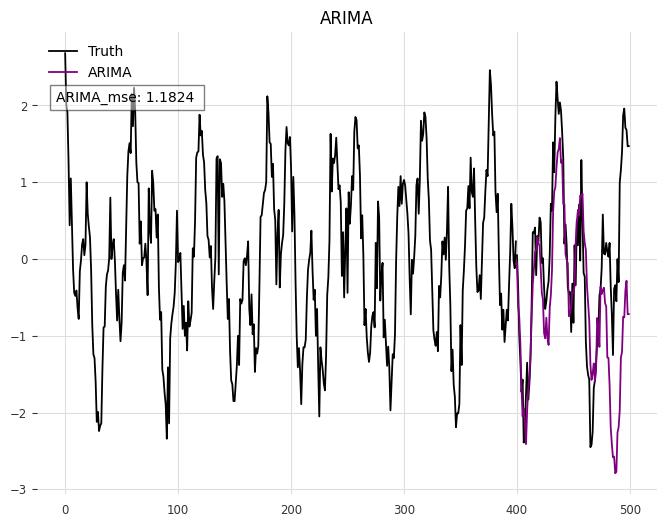}}\\
\subfloat[$\sigma=0.4$ (LLMTIME)]{
    \includegraphics[width=0.45\linewidth]{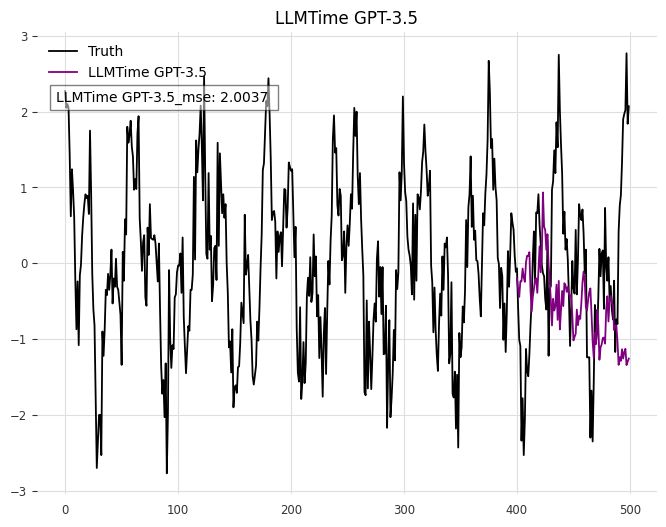}}
\subfloat[$\sigma=0.4$ (ARIMA)]{
    \includegraphics[width=0.45\linewidth]{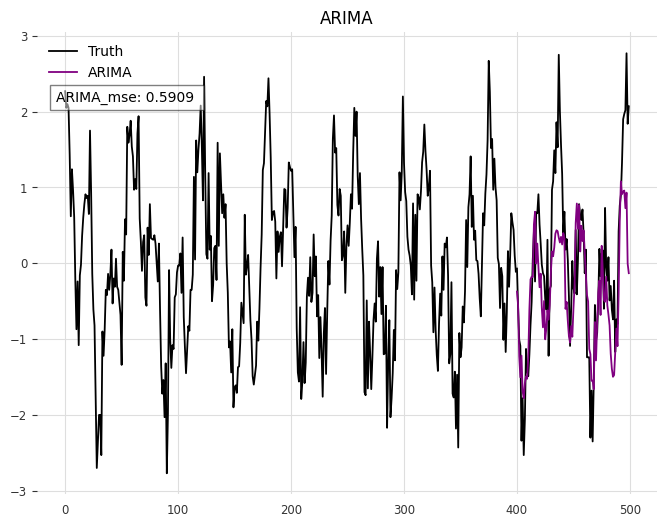}}\\

\caption{Experimental results of $cos(2\pi t)+cos(2t)+noise$.}
\label{fig11} 
\end{figure}

% \subsection{Detailed experimental results of Synthetic dataset}\label{D}

\begin{figure} [ht]
\centering
\subfloat[$\sigma=0$ (LLMTIME)]{
    \includegraphics[width=0.45\linewidth]{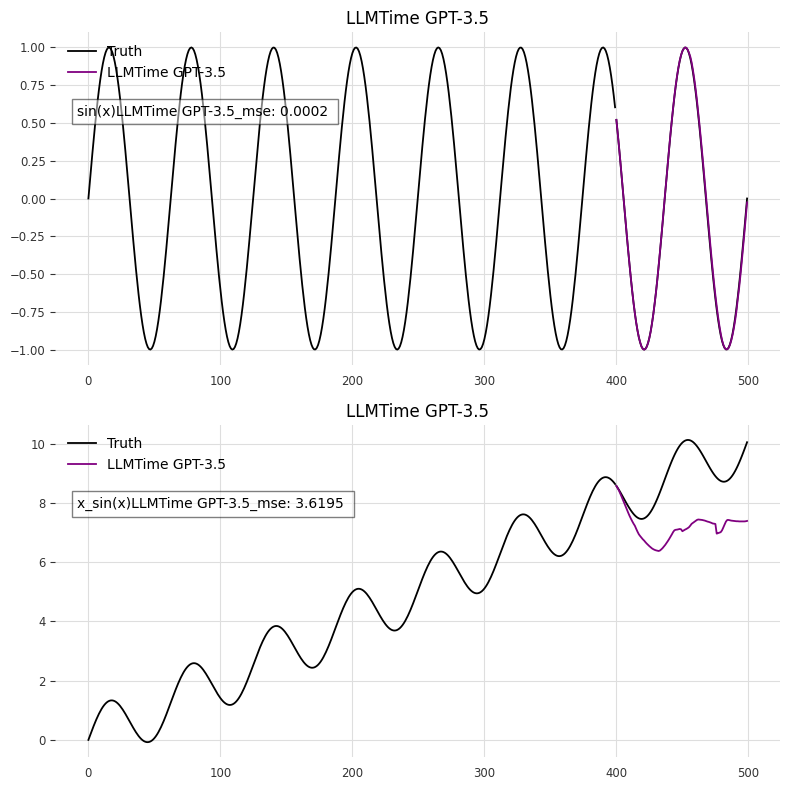}}
\subfloat[$\sigma=0$ (ARIMA)]{
    \includegraphics[width=0.45\linewidth]{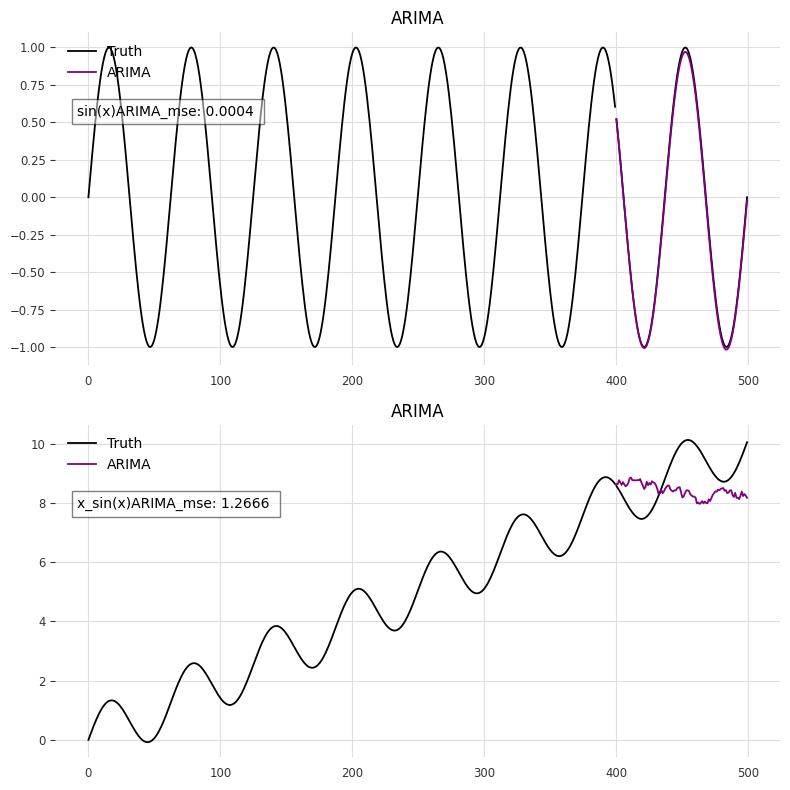}}\\
\subfloat[$\sigma=0.05$ (LLMTIME)]{
    \includegraphics[width=0.45\linewidth]{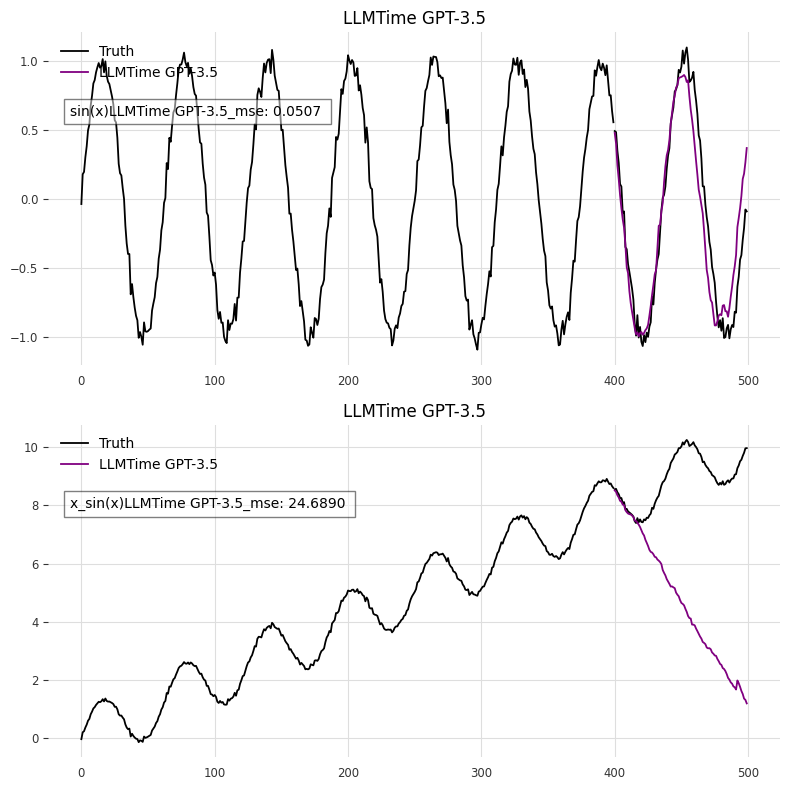}}
\subfloat[$\sigma=0.05$ (ARIMA)]{
    \includegraphics[width=0.45\linewidth]{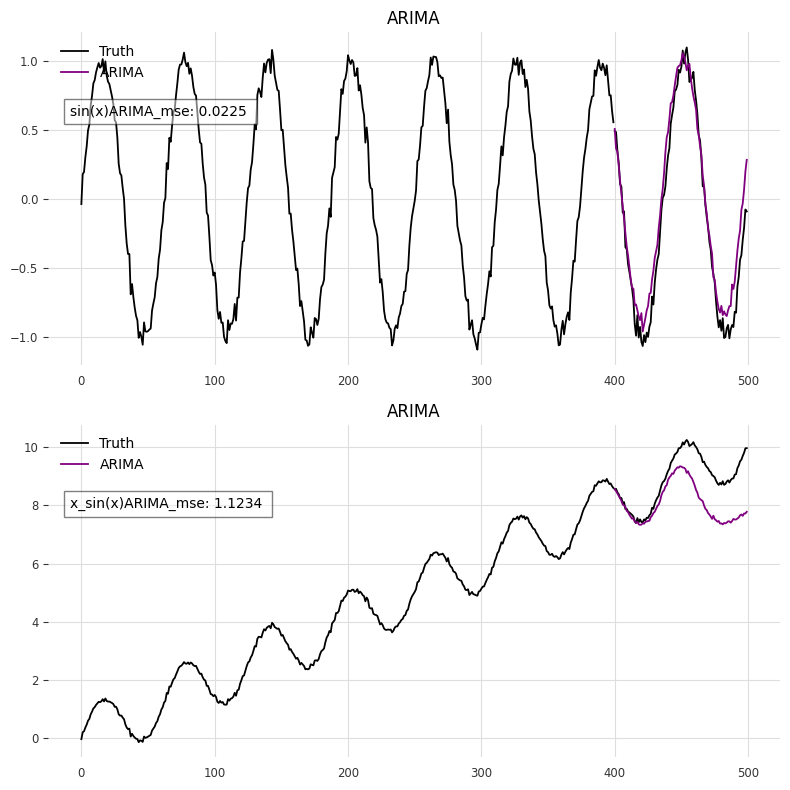}}\\
\subfloat[$\sigma=0.1$ (LLMTIME)]{
    \includegraphics[width=0.45\linewidth]{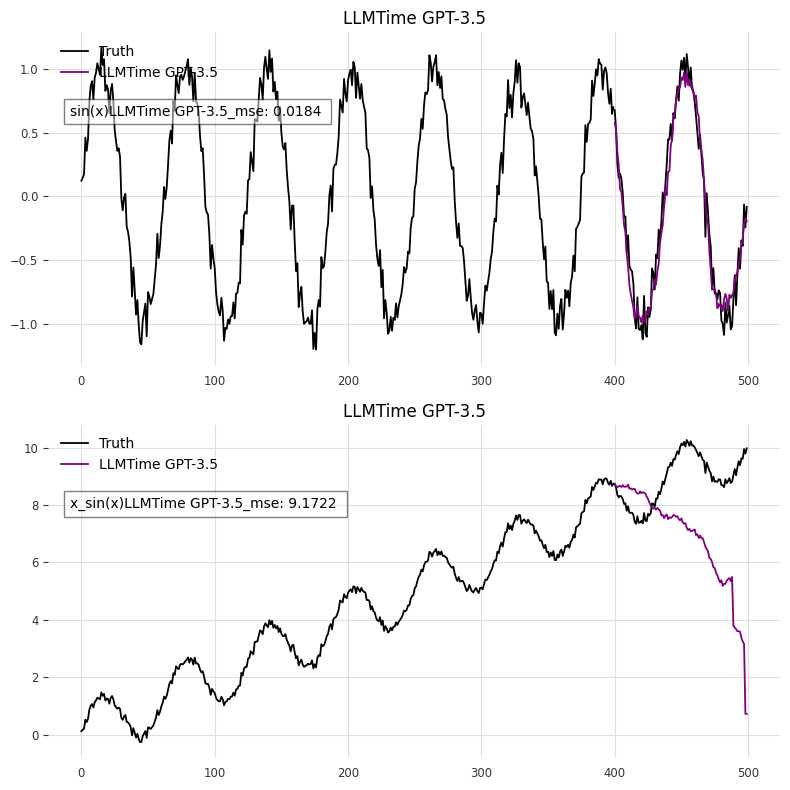}}
\subfloat[$\sigma=0.1$ (ARIMA)]{
    \includegraphics[width=0.45\linewidth]{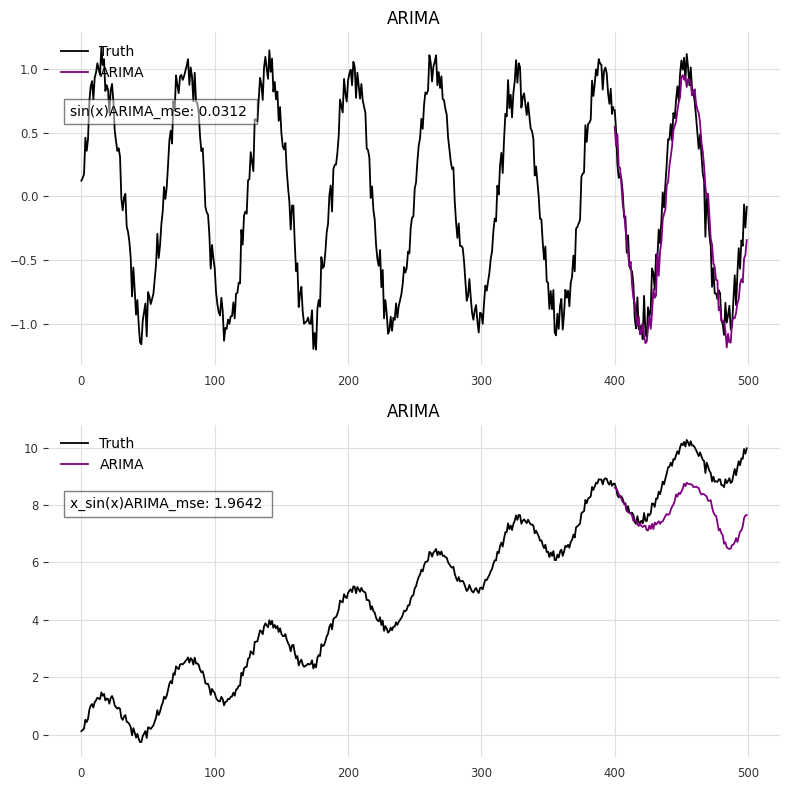}}\\
\subfloat[$\sigma=0.2$ (LLMTIME)]{
    \includegraphics[width=0.45\linewidth]{sine/sine_LLMTIME_2.png}}
\subfloat[$\sigma=0.2$ (ARIMA)]{
    \includegraphics[width=0.45\linewidth]{sine/sine_ARIMA_2.png}}\\

\caption{Experimental results of $f(t) = sin(t)+noise$ and $f(t) = 0.2t +  sin(t)+noise$.}
\label{fig13} 
\end{figure}

\end{document}